\documentclass[journal]{IEEEtran}
\ifCLASSINFOpdf
\else
\fi
\usepackage{times}
\usepackage{soul}
\usepackage{url}
\usepackage[hidelinks]{hyperref}
\usepackage[utf8]{inputenc}
\usepackage[small]{caption}
\usepackage{graphicx}
\usepackage{amsmath}
\usepackage{booktabs}
\usepackage{algorithm}
\usepackage{algorithmic}
\usepackage{multirow}
\usepackage{multicol}
\usepackage{subfigure}
\usepackage{epstopdf}
\usepackage{bbm}
\usepackage{amsmath,breqn}
\usepackage[justification=justified]{caption}

\begin{document}
%
\title{Multiple Riemannian Manifold-valued Descriptors based Image Set Classification with \\ Multi-Kernel Metric Learning}
%
%
%

\author{Rui~Wang,
        Xiao-Jun~Wu*,
        and~Josef~Kittler,~\IEEEmembership{Life~Member,~IEEE}
\thanks{The paper is supported by the National Natural Science Foundation of China (Grant No.U1836218, 61672265), UK EPSRC Grant EP/N007743/1, MURI/EPSRC/dstl Grant EP/R018456/1, and the 111 Project of Ministry of Education of China (Grant No. B12018).}
\thanks{R. Wang and X.-J. Wu are with the School of Internet of Things Engineering, Jiangnan University, Wuxi 214122, China. R. Wang and X.-J. Wu are also with Jiangsu Provincial Engineering Laboratory of Pattern Recognition and Computational Intelligence, Jiangnan University, Wuxi 214122, China e-mail: (cs\underline { }wr@jiangnan.edu.cn; xiaojun\underline { }wu\underline { }jnu@163.com).}
\thanks{J. Kittler is with the Centre for Vision, Speech and Signal Processing,
University of Surrey, Guildford GU2 7XH, U.K. e-mail: (j.kittler@surrey.ac.uk).}
}

%
%

\markboth{Journal of \LaTeX\ Class Files.}%
{Shell \MakeLowercase{\textit{et al.}}: Bare Demo of IEEEtran.cls for IEEE Journals}
%



\maketitle

\begin{abstract}
The importance of wild video based image set recognition is becoming monotonically increasing due to the large amount of video resources obtained by diversified video collection approaches, like surveillance, drive recorders, smart phones, and internet. However, the contents of these collected videos are often complicated, and how to efficiently perform set modeling and feature extraction is a big challenge for set-based classification algorithms. In recent years, some proposed image set classification methods have made a considerable advance by modeling the original image set with covariance matrix, linear subspace, or Gaussian distribution. Moreover, the distinctive geometry spanned by them are Symmetric Positive Definite (SPD) manifold, Grassmann manifold, and Gaussian embedded Riemannian manifold, respectively. As a matter of fact, most of them just adopt a single geometric model to describe each given image set, which may lose some other useful information for classification. To tackle this problem, we propose a novel algorithm to model each image set from a multi-geometric perspective. Specifically, the covariance matrix, linear subspace, and Gaussian distribution are applied for set representation simultaneously. In order to fuse these multiple heterogeneous Riemannian manifold-valued features, the well-equipped Riemannian kernel functions are first utilized to map them into high dimensional Hilbert spaces. Then, a multi-kernel metric learning framework is devised to embed the learned hybrid kernels into a lower dimensional common subspace for classification. We conduct experiments on four widely used datasets corresponding to four different classification tasks: video-based face recognition, set-based object categorization, video-based emotion recognition, and dynamic scene classification, to evaluate the classification performance of the proposed algorithm. Extensive experimental results justify its superiority over the state-of-the-art.
\end{abstract}


%
\IEEEpeerreviewmaketitle

\section{Introduction}
%
%
%
%
\IEEEPARstart{I}{n} the domain of pattern recognition and computer vision, more and more information is presented to us in the form of enormous amount of captured videos, such as surveillance videos, handheld camera videos, and internet videos, etc. One of the representative application branches of them is face recognition problem, and traditional recognition methods where the decision is based on single-shot images are achieved impressive success under restrict conditions \cite{wr1,wr2,wr3,wr4,cz,yd}. By considering each video sequence as an image set, image-set based object classification problems have recently been attracting increasing attention \cite{wr5, wr6, wr7, wr8, wr9, wr10, wr11} and exhibit extensive potential applications, including video-based face recognition \cite{wr12}, object categorization \cite{wr13}, and action recognition \cite{wr14}, etc. This is mainly because image set can provide more useful information of data variability for more robust video scene parsing under more realistic conditions. 

Different from the single-shot image based classification problem, for image set
classification, the training and testing samples are image sets, and each set generally contains a number of image instances that belong to the same category. Actually, the subject's appearance information of intra-set and/or inter-set are very likely to exhibit large variations owing to the wide range of rigid and non-rigid deformations, illumination changes, as well as different shooting conditions. Therefore, the distribution formed by the set data is often nonlinear and thus pose a key issue of how to faithfully characterize the real structure of them for classification. 

As a matter of fact, nonlinear data are often encountered in Euclidean geometry based classification tasks, which include covariance descriptors \cite{wr12,wr15,wr16}, orthogonal linear subspaces \cite{wr17}, and Gaussian distributions \cite{wr8}. However, the spaces where such nonlinear data reside on are not a vector space structure but instead a Riemannian geometry. Specifically, they are SPD manifold, Grassmann manifold, and Gaussian embedded Riemannian manifold, respectively. Hence, applying the conventional Euclidean learning techniques to the manifold-valued data straightforward is unreasonable and often leads to poor performance [15]. As a countermeasure, \cite{wr18,wr19,wr20,wr21} advocated some metrics that were designed for Riemannian manifold, incluing Affine-Invariant Riemannian metric (AIRM) \cite{wr18}, Log-Euclidean metric (LEM) \cite{wr19}, Stein divergence \cite{wr20}, and Projection metric (PM) \cite{wr21}. By utilizing these well-studied Riemannian metrics, some Euclidean learning algorithms can be generalized to Riemannian manifold by the following strategies.   

The first is to learn a Euclidean feature representation of the original Riemannian manifold-valued data by mapping the Riemannian manifold into a flat space which is an approximate Euclidean space \cite{wr13,wr16}. An alternative strategy is to embed it into a high dimensional Hilbert Space via positive definite Riemannian kernels \cite{wr10,wr12, wr22,wr23}. Compared with the former, this approach makes some Euclidean methods valid in a generalized Euclidean space, while getting a richer feature representation simultaneously. In some aspects, this approach shows better classification performance than the first one \cite{wr22,wr23}. However, the Riemannian computing methods mentioned above are actually conveying the idea of approximate computation and ignore the geometry of Riemannian manifold up to a point. To handle this problem, some Riemannian manifold dimensionality reduction methods \cite{wr14,wr21} have been suggested to directly perform a mapping from the original high-dimensional Riemannian manifold to a lower-dimensional, more discriminative one. The advantage of this type of method is that the intrinsic manifold geometry of the data has been fully considered, but it also has an inherent problem that the linear mapping function is learned on the non-linear manifold, which inevitably leads to sub-optimal results. 

In parallel with the above developments, deep neural network has become a vital tool in artificial intelligence and pattern recognition. Its advantages stem both from the ability to extract powerful feature representation and from the effective non-linear training procedure based backpropagation. Inspired by these merits, some authors try to develop the idea of conventional deep learning to Riemannian manifold, and a slice of corresponding architectures \cite{wr24,wr25,wr26} have been put forward to conduct dimensionality reduction and deep feature learning directly on Riemannian manifold. On the tail of them, the Euclidean learning methods can be applied for further computing. Undoubtedly, the introduced Riemannian manifold deep learning strategy has made significant improvements in the classification performance, which mainly owes to two reasons: 1) non-linear learning mechanism; 2) Riemannian matrix backpropagation computing.        

The above mentioned approaches for image set classification are based on the Riemannian manifold, and the distribution based methods, e.g. Single Gaussian Model (SGM) \cite{wr27}, and Gaussian mixture models (GMM) \cite{wr5,wr8}, also seems to be a favorable choice to capture the variations in the given image set. Theoretically, after the given image sets are modeled by the distribution based statistics, the similarity between any two image sets can be replaced by using Kullback-Leibler Divergence (KLD) \cite{wr28} for measurement. However, this distance metric is lack of growing discriminability for some complicated video based classification tasks \cite{wr8}. On the basis of information geometry, \cite{wr29} and \cite{wr30} point out that the space formed by $d$-dimensional Gaussian distributions can be embedded into another Riemannian manifold, specifically a SPD manifold that spanned by a set of  $(d+1)$-dimensional symmetric positive definite matrices. Therefore, the above problem can be well addressed by applying the Riemannian metric based learning algorithms \cite{wr31,wr8}.

In fact, when there exist large and complex data variations within a collected video sequence, no matter which Riemannian manifold-valued descriptor (covariance matrix, linear subspace, or Gaussian distribution), we applied for set data characterization, the discriminative information which can be provided is finite \cite{wr21,wr24,wr31,wr14}. The fundamental reason is each descriptor can only model the set data from one side of the coin. To tackle the above problem, in this paper, we propose a novel multi-kernel metric learning framework which not only can describe the original image set from a multi-geometric perspective but also can combine them for improved classification. Specifically, given an image set, we encode it by utilizing the covariance matrix, linear subspace, and Gaussian distribution simultaneously for obtaining complementary features. Since the $d$-dimensional nonsingular covariance matrix lying in a SPD manifold $Sym_d^+$, the $q$-dimensional linear subspace residing on Grassmann manifold $\mathcal{G}(q,d)$, and the $d$-dimensional Gaussian distributions can be embedded into another Riemannian manifold $Sym_{d+1}^+$, it is not trivial to fuse different topologies. We first adopt the well-studied Riemannian kernels to map each corresponding Riemannian manifold into a high dimensional Reproducing Kernel Hilbert Space (RKHS). Then, a multi-kernel metric learning algorithm is developed to fuse the learned hybrid kernels into a lower dimensional, more discriminative unified subspace for classification. Extensive classification results achieved on four widely used image set datasets validate the efficacy of the proposed method. In this paper, our main contributions can be summarized as follows:

\begin{itemize}
\item To extract complementary feature representations of image set for improved classification, we first encode each given image set (one video sequence) by three commonly used Riemannian manifold-valued descriptors, i.e., covariance matrix, linear subspace, and Gaussian distribution simultaneously.  
\item Due to the heterogeneity of the spaces spanned by these three descriptors, three well-equipped Riemannian kernel functions are then exploited to map their corresponding Riemannian manifold-valued features into high dimensional Reproducing Kernel Hilbert Spaces (RKHS) for the sake of facilitating the subsequent fusion operation.
\item Finally, we develop a multi-kernel metric learning framework to merge the generated hybrid kernel features into a lower dimensional unified subspace by jointly learning an adaptive discriminative distance metric and an adaptive weight for each local region of the produced kernel spaces. Consequently, the inter-class dispersion and intra-class compactness of the generated subspace features is enhanced.
\end{itemize}

\section{Related work}
In image set classification, the covariance matrix, linear subspace and Gaussian distribution are three commonly used Riemannian manifold-valued descriptors for image set description. For covariance matrix, its advantages are the simplicity and flexibility to capture the variations within the set \cite{wr12,wr14,wr32}, and for linear subspace, its preponderance stem both from the lower computational cost and from the ability to accommodate the effects of various intra-set variations \cite{wr10,wr21}. In comparison, the strength of Gaussian distribution is that it can describe the set data variations by estimating their first-order statistics and second order statistics simultaneously \cite{wr31,wr8}. The increasing attention and promotion of these three descriptors based image set classification problems manifests in three main factors, which are presented as follows.  

\bfseries Kernel Based Image Set Classification: \mdseries For this approach \cite{wr10,wr8,wr22,wr12,wr33}, the Riemannian manifold is embedded into a high dimensional Hilbert space via well-studied Riemannian kernel functions, which makes the Euclidean methods are easily applied for further computation. Therein, Wang \emph{et al}. \cite{wr12} employ Log-Euclidean metric (LEM) \cite{wr19} based Riemannian kernel to embed the data from SPD manifold to a generalized Euclidean space. The Kernel Discriminant Analysis (KDA) \cite{wr34} is then applied to learn a discriminative subspace for classification. Similarly, Huang \emph{et al}. \cite{wr10} map the basic elements of Grassmann manifold into a flat space by using PM based Riemannian kernel function, and try to learn a discriminant function with KDA. In order to develop the kernel based methods on Gaussian distribution, Wang \emph{et al}. \cite{wr8} investigate a series of probabilistic kernels to encode the Riemannian geometry of Gaussian distributions, and the generated kernel space is further reduced to a discriminative lower-dimensional subspace via the devised weighted KDA algorithm. However, it is obviously perceive that the learning process of such approach ignores the intrinsic Riemannian geometry of the data. 

\bfseries Manifold Dimensionality Reduction Based Image Set Classification: \mdseries To circumvent the above problem, some algorithms that jointly perform linear mapping and metric learning directly on the original Riemannian manifold have been suggested recently \cite{wr14,wr21,wr13}, and therefore a discriminative lower-dimensional one can be yielded. Harandi \emph{et al}. \cite{wr14} produce a lower-dimensional SPD manifold with an orthogonal mapping obtained by devising a discriminative metric learning framework with respect to the original high-dimensional data. To simplify the computational complexity, Huang \emph{et al}. \cite{wr13} put forward a novel Log-Euclidean metric learning algorithm to form a desirable SPD manifold by directly embedding the tangent space of original SPD manifold into a lower-dimensional one. Similarly, Huang \emph{et al}. \cite{wr21} try to learn a lower-dimensional and more discriminative Grassmannian-valued feature representations for the original high dimensional Grassmann manifold under a devised projection metric learning framework. Thanks to the advantage of fully considering the manifold geometry, the above algorithms show good classification performance. Yet, they also have an inherent design flaw, that is the mapping which is defined and learned on the non-linear Riemannian geometry is linear, which seems to be unreasonable. 

\bfseries Riemannian Deep Learning based Image Set Classification: \mdseries  As is well known, how to effectively measure the similarity between image sets is an open and challenging question, and the above mentioned Riemannian manifold learning algorithms provide some constructive ideas to address this problem. Inspired by the proven effectiveness of deep neural networks, Sun \emph{et al}. \cite{wr35} aggregate the local match kernels with a deep neural architecture to generate a global deep match kernel for similarity measurement. To take discriminative feature representation into account, Lu \emph{et al}. \cite{wr36} investigate to extract nonlinear discriminative class specific information for image set classification by integrating manifold metric learning into CNN. However, there still remains a research gap to extract more desirable feature representations for complicated classification tasks. More recently, some researchers extend the ideology of deep learning to Riemannian manifold, and raise some manifold deep learning networks to close the gap. Wherein, Huang and Val Gool \cite{wr24} first design a slice of spectral layers to deeply extract appropriate feature representations on the SPD manifold, and then propose a Riemannian matrix backpropagation algorithm for model optimization. Meanwhile, a Grassmannian deep learning architecture \cite{wr25} is devised to learn deeply selected Grassmannian-valued features. Since a specific Riemannian manifold corresponds to a specific deep learning architecture, such approach has weak of versatility and scalability.

\bfseries Multi-statistical features based Image Set Classification: \mdseries  To properly represent the image sets, the above introduced algorithms make some prior assumptions such as Gaussian distribution, linear subspace or covariance matrix. However, there often exhibit large intra-class ambiguity in wild videos, which may make these assumptions lose some hinge information for classification. To handle this problem, Lu \emph{et al}. \cite{wr37} first extract multiple order statistics of each given image set with mean, covariance matrix and tensor for set modeling, and then design a localized multi-kernel metric learning framework to perform discriminative feature fusion. However, the authors adopt one kernel function that derived from the Euclidean metric to the heterogeneous features, which may lose some capability to preserve the original geometry structure of the set data. In contrast to \cite{wr37}, Huang \emph{et al}. \cite{wr31} encode each given image set simultaneously with mean, covariance matrix and Gaussian distribution. Since different statistics span different topologies, the authors first exploit three different Riemannian kernel functions to embed these heterogeneous features into RKHS. Next, a hybrid Euclidean-and-Riemannian metric learning framework is proposed to fuse them for face recognition by learning multiple distance metrics. Whereas, it can be explicitly found that all the local regions in the learned kernel spaces share the same weights, but the importance of them is actually different. 

\section{Background Theory}
This section presents a brief introduction of the geometry of SPD manifold and Grassmann manifold, as well as of Gaussian distribution, which provides the fundamental theory for the proposed algorithm.

\subsection{The Geometry of SPD Manifold}
For all non-zero ${v} \in \mathbbm{R}^d$, a real SPD matrix ${C} \in \mathbbm{R}^{d \times d}$ has an intrinsic property, which is ${v^T C v} > 0$. The space spanned by a set of $d\times d$ SPD matrices is the interior of a convex cone in the $d(d+1)/2-$dimensional Euclidean space, signified as $Sym_d^+$. As studied in \cite{wr19,wr38}, when endowing it with an appropriate Riemannian metric, a specific Riemannian manifold can be formed, i.e., SPD manifold. Due to the topological space of SPD manifold is locally comply with Euclidean properties, the derivatives of the curves at point $C_1$ on the SPD manifold can be possibly defined under the logarithm map, which can be expressed as ${\rm log}_{C_1} : Sym_d^+ \to T_{C_1}Sym_d^+$. Therein, $T_{C_1}Sym_d^+$ denotes the tangent space of the SPD manifold at $C_1$, and a group of its adhere inner products $\langle ,\rangle_{C_1}$ is regarded as the Riemannian metric. Specifically, for any two tangent elements $T_1,T_2$, their scalar product in $T_{C}Sym_d^+$ is formulated as:
\begin{equation}
\langle T_1,T_2 \rangle_C = \langle D_C{\rm log}_{\cdot}T_1, D_C{\rm log}_{\cdot}T_2 \rangle.
\end{equation}
where $D_C{\rm log}_{\cdot}T$ is the directional derivative of the matrix logarithm at $C$ along $T$. For the logarithmic map that related to the Riemannian metric, it can be defined in terms of matrix logarithms:
\begin{equation}
{\rm log}_{C_1}(C_2) = D_{{\rm log}(C_1)}{\rm exp}_{\cdot}({\rm log}(C_2) - {\rm log}(C_1)).
\end{equation}
where $D_{{\rm log}(C)}{\rm exp}_{\cdot} = (D_C{\rm log_{\cdot}})^{-1}$, and ${\rm exp}_{\cdot}(\ )$ represents the matrix exponential map with a definition as:
\begin{equation}
{\rm exp}_{C_1}(T_2) = {\rm exp}({\rm log}(C_1) + D_{C_1}{\rm log_{\cdot}}T_2).
\end{equation}
More details about the two maps, please kindly refer to \cite{wr19,wr39}.

According to Eq.1, Eq.2 and Eq.3, the Riemannian metric on the SPD manifold can be formed by:   
\begin{equation}
\begin{split}
\mathcal{D} = \langle {\rm log}_{C_1}(C_2),{\rm log}_{C_1}(C_2) \rangle_{C_1}= || {\rm log}(C_1) - {\rm log}(C_2) ||_{F}^2.
\end{split}
\end{equation}
This metric is widely used to measure the geodesic distance between any two points on the SPD manifold named Log-Euclidean Metric(LEM) \cite{wr19}. As a result, when endowed with this Riemannian metric, the space of SPD matrices is transformed into a tangent space $T_{C}Sym_d^+$ \cite{wr19} and a valid Riemannian kernel \cite{wr12} on $Sym_d^+$ can therefore be derived by computing the inner product as:
\begin{equation}
k_{log}(C_1,C_2) = tr[{\rm log}(C_1) \bullet {\rm log}(C_2)].
\end{equation}


\subsection{The Geometry of Grassmann Manifold}
Given an orthogonal matrix $Q$ of size $d \times d$, its equivalence class $[Q]$ can be expressed as follows,
\begin{equation}
[Q] = \left\{ Q
\left(
\begin{array}{cc} 
Q_q & 0 \\ 
0 & Q_{d-q}  
\end{array}
\right) : Q_q \in O_q, Q_{d-q} \in O_{d-q} 
\right\}
\end{equation}
whose leading $q$ columns form the same subspace as those of $Q$. Here, $O_n$ is an orthogonal group composed of $d \times d$ orthogonal matrices. Actually, the equivalence class $[Q]$ represents a point lying in the Grassmann manifold $\mathcal{G}(q,d) = O_d / (O_q \times O_{d-q})$. In other words, a Grassmann manifold $\mathcal{G}(q,d)$ is spanned by a set of $q$-dimensional linear subspaces of $\mathbbm {R}^{d \times q}$. For each linear subspace, which is constituted by an orthonormal basis matrix $Y$ of size $d \times q$ ($Y^TY=I_q$, and $I_q$ is an identity matrix of size $q \times q$) is known as an element of $\mathcal{G}(q,d)$.

As well-studied in \cite{wr21,wr40}, each point in Grassmann manifold corresponding to a unique projection matrix $YY^T$ of size $d \times d$ with rank-$q$. As a result, a natural choice of inner product can be yielded under the projection operator $\Phi(Y)$, which is $\langle Y_1,Y_2 \rangle_{\Phi} = tr(\Phi(Y_1)^T,\Phi(Y_2))$, and then a geodesic distance measurement named Projection Metric is induced \cite{wr10}.
\begin{equation}
d_p(Y_1Y_1^T,Y_2Y_2^T) = 2^{-1/2}||Y_1Y_1^T - Y_2Y_2^T||_F.
\end{equation}
Since the projection mapping is continuous and differentiable, a flat space associated with the Grassmann manifold can be generated by endowing with this Riemannian metric. By computing the inner product in this flat space, we can obtain a Grassmann kernel \cite{wr10}.
\begin{equation}
k_p(Y_1Y_1^T,Y_2Y_2^T) = tr[(Y_1Y_1^T)(Y_2Y_2^T)] = ||Y_{1}^TY_2||_{F}^2.
\end{equation}
Its validity has been proven in \cite{wr10}. Please refer to \cite{wr40} for concrete mathematical theory about the Grassmannian geometry.

\subsection{The Geometry of Gaussian Distribution}
Given an image set, its mean and variations can be simultaneously captured to determine a particular Gaussian distribution, which is typically a commonly used probability distribution developed in probability theory. Therefore, the distribution of each image set can be modeled as a Single Gaussian Distribution (SGM) by estimating mean $\widetilde{m}$ and covariance matrix $\widetilde{C}$,
\begin{equation}
G(S) = \mathcal{N}(S|\widetilde{m},\widetilde{C}).
\end{equation}
where $S$ is a given image set, and the Expectation-Maximization (EM) algorithm is often exploited for estimation.

As studied in \cite{wr5}, the underlying characteristics of a family of Gaussian distributions is actually a space of constant negative curvature. Hence, it is not appropriate to endow the SGM with some Euclidean computations. Learning from the information geometry \cite{wr29,wr30}, we can see the space of Gaussian distribution can be embedded into a Riemannian manifold $Sym_{d+1}^+$. To be specific, in the information geometry, if a given random vector $v$ conforms to $\mathcal{N}(0,I)$, then its affine transformation $Av+\widetilde{m}$ conforms to $\mathcal{N}(\widetilde{m},\widetilde{C})$, and vice versa. For the covariance matrix $\widetilde{C}$, it can be decomposed into $\widetilde{C}=AA^T$, where $|A|>0$. Therefore, such a Gaussian distribution $\mathcal{N}(\widetilde{m},\widetilde{C})$ can be denoted by the affine transformation $(\widetilde{m},A)$. Based on the information geometry theory \cite{wr30}, in the Gaussian embedded space $Sym_{d+1}^+$, a $(d+1)$-dimensional SPD matrix $P$ can be uniquely represents a $d$-dimensional Gaussian model $\mathcal{N}(\widetilde{m},\widetilde{C})$ as.
\begin{equation}
\mathcal{N}(\widetilde{m},\widetilde{C}) \sim P=|A|^{-2/(d+1)}
\left(
\begin{array}{cc} 
AA^T+\widetilde{m}\widetilde{m}^T & \widetilde{m}\\ 
\widetilde{m}^T & 1  
\end{array}
\right).
\end{equation}
For a more detailed introduction to the Gaussian embedding process, please refer to \cite{wr30}.   

Since we embed such a $d$-dimensional Single Gaussian Model into another SPD manifold $Sym_{d+1}^+$, the well-studied LEM can be applied to replace the KL divergence to measure the distance between two probability distributions. Moreover, as studied in \cite{wr31}, we can formulate a corresponding kernel of Gaussian distributions as:
\begin{equation}
k_G(G_1,G_2) = tr(\rm log(P_1) \bullet \rm log(P_2)).
\end{equation}
where $P_1,P_2$ represent the $(d+1)$-dimensional SPD matrices that corresponding to two Gaussian models.

\section{Proposed method}

Fig.1 uses a schematic diagram to intuitively present the proposed method. For each given image set, as discussed before, three different Riemannian manifold-valued descriptors are adopted to simultaneously model it for the purpose of extracting complementary feature information. Due to the spaces spanned by them are three types of Riemannian manifolds: $Sym_d^+$, $Sym_{d+1}^+$ and $\mathcal{G}(q,d)$, then how to fuse these heterogeneous features is becoming an essential problem for classification. To this end, a multi-kernel metric learning framework is designed in the hybrid kernel spaces produced by three well-equipped kernel functions. Eventually, a more discriminative common subspace can be yielded. 

\begin{figure*}[!t]
 \centering
 \includegraphics[scale=0.3]{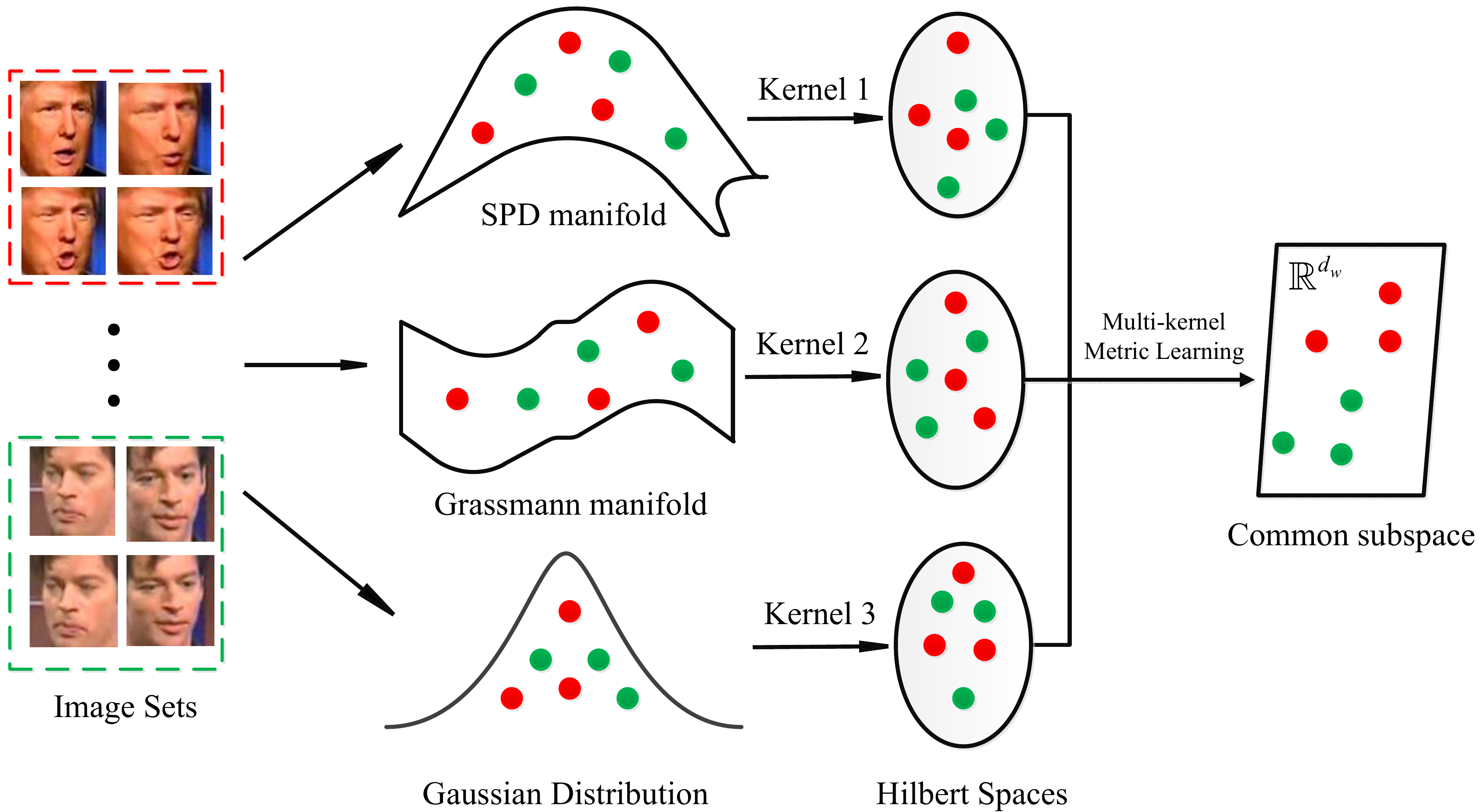}
 \caption{Schematic diagram of the proposed image set classification framework. The first column in this figure represents the given image sets. Then, we simultaneously use three Riemannian manifold-valued descriptors (i.e., covariance matrix, linear subspace and Gaussian distribution) to encode each image set, which reside in SPD manifold $Sym_{d}^+$, Grassmann manifold $\mathcal{G}(q,d)$ and Gaussian embedded Riemannian manifold $Sym_{d+1}^+$ respectively (the second column). Next, three well-equipped Riemannian kernels are applied to map such heterogeneous spaces into high dimensional Hilbert spaces (the third column). Finally, a multi-kernel metric learning framework is designed to fuse these hybrid and complementary kernel features into a lower-dimensional, more discriminative common subspace for classification.}
 \label{fig:label}
\end{figure*}

\subsection{Set Modeling with Multiple Riemannian manifold-valued descriptors}
Let $S_i=[s_1,s_2,...,s_{n_i}]$ be the $i$-th given image set with $n_i$ entities, where $s_i \in \mathbbm{R}^{d\times1}$ represents the $i$-th image sample. Given an image set, we encode it with the following three descriptors, and the extracted complementary feature information can be regarded as the new image set features for the subsequent computations.
\begin{itemize}
\item \bfseries Set Modeling with Covariance Matrix: \mdseries The second-order statistics is a widely used model for set representation. Its advantages are the simplicity and flexibility to model the variations within the set with no assumption about the set data distribution. For $S_i$, we can compute its covariance matrix as:
\begin{equation}
C_i=\frac{1}{n_i-1}\sum_{i=1}^{n_i}(s_i-m)(s_i-m)^T.
\end{equation}
where $m$ is the mean of $S_i$. As studied in \cite{wr19,wr38}, the space spanned by positive definite covariance matrices is SPD manifold $Sym_{d}^+$. Hence, we apply the following tactic to maintain the positive definiteness of $C_i$.
\begin{equation}
C_i \gets C_i+\frac{tr(C_i)}{\alpha}I_d.
\end{equation}
where $I_d$ is an identity matrix of size $d \times d$, and we set $\alpha=10^3$ in all the experiments.
\item \bfseries Set Modeling with Linear Subspace: \mdseries For linear subspace, it can be regarded as the subspace-based statistics, which has the advantages of lower computational complexity and accommodating to the effects of various within-set variations. For $S_i$, its $q$-dimensional linear subspace used for set representation is formed by an orthonormal basis matrix $Y_i \in \mathbbm{R}^{d \times q}$, which can be easily obtained by:
\begin{equation}
S_iS_{i}^T \simeq Y_i\Lambda_iY_{i}^T.
\end{equation}
where $\Lambda_i$ and $Y_i$ respectively represent the matrices of $q$ largest eigenvalues and their corresponding $q$ largest eigenvectors. As studied in \cite{wr40}, the linear subspace resides on Grassmann manifold.
\item \bfseries Set Modeling with Gaussian Distribution: \mdseries Gaussian distribution is actually a probability distribution developed in probability theory. It is often utilized to model the set data by simultaneously capturing its first-order statistics and second-order statistics under EM algorithm. Therefore, when specifying $(\widetilde{m_i},\widetilde{C_i})$, a Single Gaussian Model (SGM) can be formulated as.
\begin{equation}
G(S_i) = \mathcal{N}(S_i|\widetilde{m_i},\widetilde{C_i}).
\end{equation}
In the information geometry \cite{wr29,wr30}, the $G(S_i)$ is typically lie in another Riemannian manifold $Sym_{d+1}^+$, which is spanned by a family of SPD matrices of size $(d+1)\times(d+1)$. Therefore, we adopt the same strategy as used in Eq.13 to keep the non-singularity of $\widetilde{C_i}$.      
\end{itemize}

\subsection{Multi-Kernel Metric Learning for Heterogeneous Features Fusion}
As stated earlier, different descriptors reside on different Riemannian manifolds, straightforwardly combining them for classification is inappropriate. In this part, we will present this multi-kernel metric learning framework designed for fusing the extracted heterogeneous but complementary feature representations in detail.

Let $T=[S_1,S_2,...,S_N]$ represents the gallery consisting of $N$ image sets, where $S_i=[s_1,s_2,...,s_{n_i}]$ is the $i$-th image set, $1 \le {i \le N}$, and $n_i$ demonstrates the number of images in this set. For each $S_i$, we use covariance matrix, linear subspace and Gaussian distribution to model it, respectively. We use $X^q=[X_{1}^q,X_{2}^q,...,X_{N}^q]$ to represent the $q$-th generated feature set of the gallery, and $X_{i}^q \in \mathbbm{R}^{d_q}$ is the $q$-th Riemannian manifold-valued feature representation extracted from $S_i$. Here, $q=1 \to Q$ and we set $Q=3$ as three different models are utilized to describe the set data. In order to aggregate such heterogeneous feature representations, three well-studied Riemannian kernel functions are then applied for high dimensional feature embeddings in the light of the proven success of kernel learning \cite{wr41, tbd1, tbd2}. This process is implemented by mapping the original Riemannian manifold-valued features into a Hilbert space, and then computing the dot product in it. We use $\phi_{i}^q$ to represent the generated new feature representation of $X_{i}^q$, and the non-linear mapping function is formulated as: $\phi \; {\rm :} \; \mathbbm{R}^{d_q} \to \mathcal{F}$, where $\mathbbm{R}^{d_q}$ denotes the produced $q$-th Riemannian space and $\mathcal{F}$ is the transformed Hilbert space. Though the mapping function $\phi$ is usually implicit, we first directly use it to formulize our method for simplicity. For classification, the first and foremost task is to measure the similarity between a given a pair of training image sets $S_i$ and $S_j$ by defining a distance metric in $\mathcal{F}$ as:
\begin{equation}
d(S_i,S_j) = tr\lbrack\sum_{q=1}^{Q}\xi_q(\phi_{i}^q)(\phi_{i}^q-\phi_{j}^q)^TU(\phi_{i}^q-\phi_{j}^q)\xi_q(\phi_{j}^q)\rbrack.
\end{equation}
where $\xi_q(\phi_{i}^q)$ is a gating model defined to assign different positive weights to different $\phi_{i}^q$, which will be introduced later, and $U$ is the Mahalanobis matrix needs to be learned. Due to its symmetric positive semi-definite property, we can look for a non-square matrix $W=[w_1,w_2,...,w_{d_w}]$ to re-represent it as $U=WW^T$, and the Eq.16 can therefore be rewritten as.       
\begin{equation}
d(S_i,S_j) = tr\lbrack W^T(\sum_{q=1}^{Q}\xi_q(\phi_{i}^q)(\phi_{i}^q-\phi_{j}^q)(\phi_{i}^q-\phi_{j}^q)^T\xi_q(\phi_{j}^q))W\rbrack
\end{equation}

Our next target is to learn the transformation matrix $W$, so that the hybrid kernel features can be mapped into a desirable unified space for more discriminative classification. To achieve this purpose, we attempt to simultaneously maximize the distance of all the between-class sample pairs and minimize the distance of all the within-class sample pairs in the gallery with the following objective function:
\begin{equation}
W^*=\mathop{\rm {arg\,max}}\limits_{W}J(W)=\mathop{\rm {arg\,max}}\limits_{W}\frac{R_b(W)}{R_w(W)}
\end{equation}
where $R_b(W),\;R_w(W)$ denote the between-class dispersion and within-class compactness, respectively, and can be formulated as:
\begin{equation}
R_w(W)=\frac{1}{N_w}\sum_{i=1}^N\sum_{j{\rm :}l_i=l_j}d(S_i,S_j)=tr(W^T\Psi_w W)
\end{equation}
\begin{equation}
R_b(W)=\frac{1}{N_b}\sum_{i=1}^N\sum_{j{\rm :}l_i \neq l_j}d(S_i,S_j)=tr(W^T\Psi_b W)
\end{equation}
where $N_w,\;N_b$ represent the number of sample pairs from the intra-class and inter-class in the training set, $l_i,\;l_j$ denote the category labels of $S_i$ and $S_j$, and $\Psi_w,\;\Psi_b$ are the intra-class scatter matrix and inter-class scatter matrix, which can be formulated as:  
\begin{equation}
\Psi_w=\frac{1}{N_w}\sum_{i=1}^N\sum_{j{\rm :}l_i=l_j}\sum_{q=1}^Q\xi_q(\phi_{i}^q)(\phi_{i}^q-\phi_{j}^q)(\phi_{i}^q-\phi_{j}^q)^T\xi_q(\phi_{j}^q))
\end{equation}
\begin{equation}
\Psi_b=\frac{1}{N_b}\sum_{i=1}^N\sum_{j{\rm :}l_i \neq l_j}\sum_{q=1}^Q\xi_q(\phi_{i}^q)(\phi_{i}^q-\phi_{j}^q)(\phi_{i}^q-\phi_{j}^q)^T\xi_q(\phi_{j}^q))
\end{equation}

Clearly, it is arduous for us to perform subsequent computations, which dues to the form of $\phi$ is unknown and thus is impossible to compute $\Psi_w$ and $\Psi_b$ in the mapped space. However, when we express the basis $w_h$ as a linear combination of all the training samples in the feature space $\mathcal{F}$, i.e.,
\begin{equation}
    w_h = \sum_{i=1}^Ne_{i}^h\phi_{i}^q
\end{equation}
where $e_{i}^h$ are the representation coefficients. As a result, the above problem can be addressed by using the kernel trick method \cite{wr42} as:
\begin{equation}
    \sum_{q=1}^Qw_{h}^T\phi_{i}^q=\sum_{i=1}^N\sum_{q=1}^Qe_{i}^h(\phi_{i}^q)^T\phi_{i}^q=\sum_{q=1}^Q(e^h)^TK_{.i}^q
\end{equation}
where $e^h\in R^{N\times1}$ is a column vector and $e_{i}^h$ is its $i$-th entry, and $K_{.i}^q$ is the $i$-th column of the $q$-th kernel matrix $K^q$. Here, $K^q$ is of size $N\times N$, generated from the $q$-th Riemannian manifold feature using the corresponding $q$-th Riemannian kernel function.

Hence, the intra-class and inter-class scatter matrices can be respectively re-expressed as.
\begin{equation}
\Upsilon_{w}=\frac{1}{N_w}\sum_{i=1}^N\sum_{j{\rm :}l_i=l_j}\sum_{q=1}^Q\xi_q(\phi_{i}^q)(K_{.i}^q-K_{.j}^q)(K_{.i}^q-K_{.j}^q)^T\xi_q(\phi_{j}^q)
\end{equation}
\begin{equation}
\Upsilon_{b}=\frac{1}{N_b}\sum_{i=1}^N\sum_{j{\rm :}l_i \neq l_j}\sum_{q=1}^Q\xi_q(\phi_{i}^q)(K_{.i}^q-K_{.j}^q)(K_{.i}^q-K_{.j}^q)^T\xi_q(\phi_{j}^q)
\end{equation}
Thus, the objective function can be rewritten as.
\begin{equation}
E^*=\mathop{\rm {arg\,max}}\limits_{E}J(E)=\mathop{\rm {arg\,max}}\limits_{E}\frac{tr(E^T\Upsilon_b E )}{tr(E^T\Upsilon_w E)}
\end{equation}

Next, we have another problem need to be discussed, which is the gating model. In fact, the specific form of the gating model is not fixed. In this study, it is defined as follow \cite{wr42}:
\begin{equation}
    \xi_q(\phi_{i}^q)=\frac{{\rm exp}(g_{q}^T\phi_{i}^q+\rho_q)}{\sum_{r=1}^Q{\rm exp}(g_{r}^T\phi_{i}^r+\rho_r)}
\end{equation}
where $g_q$ and $\rho_q$ are the parameters of this gating model. It is interesting to find this gating model is increased incrementally with the importance of $\phi_{i}^q$, and the softmax can guarantee its nonnegativity. However, this gating model is currently difficult to play a part for the reason of the implicit form of $\phi$. To tackle this, we employ the similar way as introduced in Eq.23.    
\begin{equation}
    g_{q}^T\phi_{i}^q=\delta_{q}^T(\phi_{i}^q)^T\phi_{i}^q=\delta_{q}^TK_{.i}^q
\end{equation}
Then, this gating model can be reformulated as,
\begin{equation}
    \xi_q(\phi_{i}^q)=\frac{{\rm exp}(\delta_{q}^TK_{.i}^q+\rho_q)}{\sum_{r=1}^Q{\rm exp}(\delta_{r}^TK_{.i}^r+\rho_r)}
\end{equation}
where $\delta \in R^{N\times1}$ and $\rho \in R^1$ are the two parameters need to be learned in this gating model. 

\subsection{Optimization}
A difficulty in solving the trace ratio problem in Eq.27 arises from the fact that a closed-form solution for the transformation matrix $E$ is unknown. This is mainly because the existed outcome interdependence between $E$ and $(\delta_q,\rho_q)$. Hence, we use an iterative mode to handle this problem. To be specific, we first fix the values of $\delta_q$ and $\rho_q$ with a randomly generated vector of size $N\times1$ and a randomly generated constant respectively, to get the new $E$, and then update $\delta_q$ and $\rho_q$ with the updated $E$, iteratively.

\bfseries 1. Computation of $E$. \mdseries  Conventionally, such nonconvex trace ratio problem in Eq.27 is often transformed into a simpler ratio trace problem \cite{wr43}, which is shown in Eq.31 to get the closed-form solution.
\begin{equation}
\begin{split}
    E^*= \mathop{\rm {arg\,max}}\limits_{E}J(E)&=\mathop{\rm {arg\,max}}\limits_{E}tr[(E^T\Upsilon_w E )^{-1}(E^T\Upsilon_bE)]\\
    &=\mathop{\rm {arg\,max}}\limits_{E}\frac{|E^T\Upsilon_bE|}{|E^T\Upsilon_wE|}
\end{split}
\end{equation}
Obviously, it can be easily solved with the Eigen-Decomposition method. However, this approximation may sacrifice the potential discriminatory ability of the produced lower-dimensional feature \cite{wr44}. Instead, we follow an efficient way \cite{wr44} to directly solve the trace ratio problem defined in Eq.27.  

Denote $\Upsilon_t=\Upsilon_w+\Upsilon_b$, and make an assumption that $E^TE=I_{d_N}$, then the trace ratio problem is equivalently changed to
\begin{equation}
    E^*= \mathop{\rm {arg\,max}}\limits_{E^TE=I_{d_N}}J(E)=\mathop{\rm{arg\,max}}\limits_{E^TE=I_{d_N}}\frac{tr(E^T\Upsilon_b E)}{tr(E^T \Upsilon_t E)}. 
\end{equation}
Without losing generality, we briefly summarize this procedure as the following two steps:

\bfseries (1) Remove the Null Space of $\Upsilon_t$. \mdseries Because of the positive semi-definite property of $\Upsilon_w$ and $\Upsilon_b$, the intersection of their null space is equal to the null space of $\Upsilon_t$. As a matter of fact, there does not exist any discriminative feature information lie in the null space of $\Upsilon_t$, so it could be removed from the solution space without losing accuracy. For $\Upsilon_t$, its singular value decomposition is expressed as:
\begin{equation}
    \Upsilon_t=A\Sigma A^T
\end{equation}
where  $\Sigma=[\lambda_1,\lambda_2,...,\lambda_{d_e}]$, and $d_e$ represents the number of positive singular values of $\Upsilon_t$. Then, the solution space is restrict to a new space formed by the column vectors of $A$, which is $E=AV$, and $V$ is of size $d_e \times d_w$. Consequently, the trace ratio problem in Eq.32 can be converted into:
\begin{equation}
    V^*=\mathop{\rm{arg\,max}}\limits_{V^TV=I_{d_w}}\frac{tr(V^T\Upsilon_{b}^a V)}{tr(V^T\Upsilon_{t}^aV)}.
\end{equation}
where $\Upsilon_{t}^a=A^T\Upsilon_tA$ and $\Upsilon_{b}^a=A^T\Upsilon_bA$. Now, $\Upsilon_{t}^a$ is positive definite.

\bfseries (2) Iterative Optimization. \mdseries Based on the trace ratio problem defined in Eq.34, we solve a trace difference problem in each step:
\begin{equation}
    V^*=\mathop{\rm{arg\,max}}\limits_{V^TV=I_{d_w}}tr[V^T(\Upsilon_{b}^a-\lambda^i\Upsilon_{t}^a)V].
\end{equation}
where $i=1,2,3...,R$, and $\lambda^i$ is the $i$-th trace ratio value computed from the transformation matrix $V^{i-1}$ of the last step and can be formulated as:
\begin{equation}
    \lambda^i=\frac{tr[(V^{i-1})^T\Upsilon_{b}^aV^{i-1}]}{tr[(V^{i-1})^T\Upsilon_{t}^aV^{i-1}]}.
\end{equation}
Here, $V^0$ is endowed with a randomly initialized columnly orthogonal matrix, and with the obtained $\lambda^i$, the trace difference problem of $i$-th step is constructed as:
\begin{equation}
    V^i=\mathop{\rm{arg\,max}}\limits_{V^TV=I_{d_w}}tr[V^T(\Upsilon_{b}^a-\lambda^i\Upsilon_{t}^a)V].
\end{equation}
By this time, the Eigen-Decomposition method is utilized to obtain the $i$-th projection matrix $V^i$. For the sake of orthogonal transformation invariance, $V^i$ is reshaped by adopting singular value decomposition to $\Upsilon_{t}^v$ as $\Upsilon_{t}^v=V^i\Sigma^i(V^i)^T$, where $\Upsilon_{t}^v=V^i(V^i)^T\Upsilon_{t}^aV^i(V^i)^T$. Then, iterating the above operations to get the desirable $d_e \times d_w$ projection matrix $V$. This algorithm is proved to be converged to the global Optimum. For a more detailed treatment, please refer to \cite{wr44}.      

\bfseries 2. Computation of $\delta_q$ and $\rho_q$. \mdseries Having computed $E$, we first conduct partial derivatives of $J(E)$ with respect to $\delta_q$ as:  
\begin{equation}
\begin{split}
    \frac{\partial J(E)}{\partial \delta_q}&=\frac{\frac{\partial H_b}{\partial \delta_q}H_w-\frac{\partial H_w}{\partial \delta_q}H_b}{(H_w+H_b)^2} \\
    &=\frac{EE^T[\frac{\partial \Upsilon_b(\delta_q,\rho_q)}{\partial \delta_q}H_w-\frac{\partial \Upsilon_w(\delta_q,\rho_q)}{\partial \delta_q}H_b]}{(H_w+H_b)^2}
\end{split}
\end{equation}
where $H_b=tr(E^T\Upsilon_b(\delta_q,\rho_q)E), H_w=tr(E^T\Upsilon_w(\delta_q,\rho_q)E)$, and the specific forms of $\frac{\partial \Upsilon_b(\delta_q,\rho_q)}{\partial \delta_q}$ and $\frac{\partial \Upsilon_w(\delta_q,\rho_q)}{\partial \delta_q}$ are respectively presented as follows:
\begin{equation}
\begin{split}
    \frac{\partial \Upsilon_b(\delta_q,\rho_q)}{\partial \delta_q}&=\frac{1}{N_b}\sum_{i=1}^N\sum_{j{\rm :}l_i \neq l_j}\sum_{k=1}^Q\xi_k(\phi_{i}^k)(K_{.i}^k-K_{.j}^k)\\
    &(K_{.i}^k-K_{.j}^k)^T\xi_k(\phi_{j}^k)[K_{.i}^q(\beta_{q}^k-\xi_q(\phi_{i}^q))+\\
    &K_{.j}^q(\beta_{q}^k-\xi_q(\phi_{j}^q))]
\end{split}
\end{equation}
\begin{equation}
\begin{split}
    \frac{\partial \Upsilon_w(\delta_q,\rho_q)}{\partial \delta_q}&=\frac{1}{N_w}\sum_{i=1}^N\sum_{j{\rm :}l_i= l_j}\sum_{k=1}^Q\xi_k(\phi_{i}^k)(K_{.i}^k-K_{.j}^k)\\
    &(K_{.i}^k-K_{.j}^k)^T\xi_k(\phi_{j}^k)[K_{.i}^q(\beta_{q}^k-\xi_q(\phi_{i}^q))+\\
    &K_{.j}^q(\beta_{q}^k-\xi_q(\phi_{j}^q))]
\end{split}
\end{equation}
where $\beta_{q}^k=1$ if $q=k$ and 0 otherwise. Similarly, we can easily get the partial derivative of $J(E)$ with respect to $\rho_q$ by referring to the above process, and we omit it here for simplicity.

After obtaining $\frac{\partial J(E)}{\partial \delta_q}$ and $\frac{\partial J(E)}{\partial \rho_q}$, the gradient ascent method is applied to train the gating model:
\begin{equation}
    \delta_{q}^{t+1}=\delta_{q}^t+\gamma\frac{\partial J(E)}{\partial \delta_q}
\end{equation}
\begin{equation}
    \rho_{q}^{t+1}=\rho_{q}^t+\gamma\frac{\partial J(E)}{\partial \rho_q}
\end{equation}
where $\gamma$ is the learning rate and configured as $10^{-4}$ in the experiments. 

Having updated $\delta_q$ and $\rho_q$, we need to utilize them to update the values of the $\xi_q(\phi_{i}^q)$, $\Upsilon_w$ and $\Upsilon_b$, respectively. So that the transformation matrix $E$ can be updated by re-solving the trace ratio problem defined in Eq.32. Afterwards, repeating this staggered process for a certain number of iterations until the conditions are satisfied. We summarize the proposed image set classification algorithm in Algorithm 1. 

\begin{table}[!ht]
\rule[0.1cm]{8.8cm}{1.0pt}
\leftline {\textbf {Algorithm 1.} Metric Learning for Heterogeneous Features Fusion}\\
\rule[0.1cm]{8.8cm}{1.0pt} \\
\textbf{Input:} Training image sets $T$, label matrix $L$, $Q$ different kernel matrices $K^q$ ($1 \le {q \le Q}$), the number of iterations $B$, target feature dimension $d_w$ and convergence error $\varepsilon$. \\
\textbf{Output:} Target transformation matrix $E$ and two parameters $\delta_q,\rho_q$. \\
\textbf{Step 1 (Initialization):} Initialize $\delta_{q}^0$ with an arbitrary column vector, and $\rho_{q}^0$ with a small random number.\\
\textbf{Step 2 (Optimization):} For $i=1,2,...,B$, repeat
\begin{enumerate}
\item Use Eq.25 and Eq.26 to compute $\Upsilon_w,\Upsilon_b$, respectively.
\item Solve the trace ratio problem defined in Eq.32 and get the $i$-th transformation matrix $E^i=[e_1,e_2,...,e_{d_w}]$.
\item Update $\delta_q$ and $\rho_q$ by using gradient ascent method:\\
$$\quad \delta_{q}^{t+1} \longleftarrow \delta_{q}^t+\gamma\frac{\partial J(E)}{\partial \delta_q}$$
$$\quad \rho_{q}^{t+1} \longleftarrow \rho_{q}^t+\gamma\frac{\partial J(E)}{\partial \rho_q}$$
\item Check convergence: \\ 
if $i>2$, $|\delta_{q}^{(i+1)}-\delta_{q}^i|<\varepsilon$ and $|\rho_{q}^{(i+1)}-\rho_{q}^i|<\varepsilon$ or  $|E^{i+1}-E^i|<\varepsilon$, turn to Step 3.
\end{enumerate}
\textbf{Step 3 (Output):} Transformation matrix $E$ and $\delta_q,\rho_q$ \\
\rule[0.1cm]{8.8cm}{1.0pt}
\end{table}

\subsection{Classification}

In the testing phase, we first apply the three different Riemannian manifold-valued descriptors to encode a given testing image set $S_{te}$, and we here use $X_{te}^q$ to represent the extracted $q$-th Riemannian feature, where $q=1 \to Q$. Then, we measure the similarity between $S_{te}$ and all the training sets in the form of three different computed kernel vectors with each denoted by $K_{.te}^q$. Afterwards, the distance between $S_{te}$ and each training set $S_i$ is formulated as follows. 
\begin{equation}
\begin{split}
    d(S_{te},S_i)&=tr\lbrack\sum_{q=1}^{Q}\xi_q(\phi_{te}^q)(K_{.te}^q-K_{.i}^q)^TEE^T\\
    &(K_{.te}^q-K_{.i}^q)\xi_q(\phi_{i}^q)\rbrack
\end{split}
\end{equation}

Lastly, we use the nearest neighbor classifier to do classification.

\subsection{Computational Complexity}
According to the Algorithm 1, we can intuitively realize that the time consumption in the training stage is mainly manifested in three aspects: 1) building $Q$ different kernel matrices, which is $\mathcal{O}(QN^2)$. Here, $N$ is the number of training sets; 2) computing the intra-class scatter matrix $\Upsilon_w$ and inter-class scatter matrix $\Upsilon_b$. Here, we use $F_i$ to represent the number of image sets that are used to train in the $i$-th category. As a result, we respectively need to pay $\mathcal{O}(NF_i)$ and $\mathcal{O}(N(N-F_i))$ for computing them; 3) updating $\delta_q$ and $\rho_q$, which is $\mathcal{O}(2BN^2))$ ($B$ is the number of the iterations). As a result, the computational complexity of the training phase is $\mathcal{O}((2B+1+Q)N^2)$. In the testing phase, the main time cost is to construct the $Q$ different similarity matrices and compute the distance between each testing image set and each training image set, respectively. Clearly, the computational complexity of them is $\mathcal{O}(QN_{te}N)$ and $\mathcal{O}(N_{te}N)$, where $N_{te}$ represents the number of testing samples. Considering that $B \ll N^2$ and $Q \ll N$, the primary time cost of the proposed algorithm is $\mathcal{O}(N^2+N_{te}N)$.

\subsection{Relation with the Previous Works}
The proposed method is similar to \cite{mmml, wr31}. We summarize some essential differences between ours and those introduced in \cite{mmml, wr31} in the following paragraphs.    
\begin{itemize}
\item \bfseries Relation with \cite{mmml}: \mdseries In fact, the proposed algorithm is an extension of our previous work \cite{mmml}. The essential differences between the proposed work and the conference paper lie in the following five aspects: 1) in addition to set modeling with covariance matrix and linear subspace in \cite{mmml}, this paper also exploits gaussian distribution to encode the original set data for the sake of mining more useful information of intra-class variations; 2) Due to the space formed by a set of gaussian distributions is another Riemannian manifold $Sym_{d+1}^+$, a well-equipped Riemannian kernel function is further applied to it for the purpose of preserving the structural information of the Riemannian manifold-valued data in the Hilbert space embedding; 3) Due to the discriminability of each local region in the produced kernel spaces is different, this paper integrates the devised multi-kernel learning algorithm into our originally proposed metric learning framework \cite{mmml} for the sake of learning an adaptive weight for each, while \cite{mmml} assigns the same weight to them; 4) To optimize the transformation matrix, this paper follows an efficient way \cite{wr44} to directly solve the trace ratio problem, while \cite{mmml} transforms this problem into a simpler ratio trace problem for approximation computing; 5) Besides the video-based face recognition and set-based object categorization tasks in the conference paper, we further assess the proposed work on video-based emotion recognition and dynamic scene classification tasks by making extensive experiments on two challenging video-based datasets: AFEW \cite{afew} and MDSD \cite{wr35}.
\item \bfseries Relation with \cite{wr31}: \mdseries The proposed method and \cite{wr31} not only focus on building reliable set models, but also focus on learning discriminative subspace feature representations. However, their essential differences are as follows: 1) besides covariance matrix and gaussian distribution which are used for set description, the proposed algorithm also makes use of linear subspace to characterize the original set data, while \cite{wr31} extracts their first-order information. The linear subspace has been proven to be able to accommodate the effects of various intra-set variations; 2) The weight corresponding to each local region of the learned kernel spaces is obtained by adaptive learning under the designed multi-kernel metric learning framework in this paper, while which are the same in \cite{wr31}; 3) For optimization, this paper first formulates the feature fusion problem into the trace ratio form, and then exploits ITR \cite{wr44} and gradient descent method to iteratively solve it, while \cite{wr31} utilizes the LogDet divergence \cite{ldd} based constraint to formulate the feature fusion problem, and adopts the cyclic Bregman projection method \cite{cbp} to solve it; 4) This paper evaluates the proposed algorithm on four different video-based classification tasks, and the extensive classification results justify its effectiveness, while \cite{wr31} concentrates on video-based face recognition task.                  

\end{itemize}

\section{Experiments}
In this section, we evaluate the proposed algorithm $\footnote{The source code will be released on: https://github.com/GitWR}$ on four image set classification tasks: video-based face recognition, set-based object categorization, video-based emotion recognition and dynamic scene classification, respectively.

\subsection{Comparative Methods and Settings}
In this paper, we compare the proposed algorithm with some representative image set classification methods which can be divided into five categories as follows:

\begin{itemize}
\item \bfseries Kernel based methods: \mdseries Grassmann Discriminant Analysis (GDA) \cite{wr10}, Grassmannian Graph-Embedding Discriminant Analysis (GEDA) \cite{wr45}, Covariance Discriminative Learning (CDL) \cite{wr12}, Riemannian Sparse Representation (RSR) \cite{wr22} and Discriminant Analysis on Riemannian manifold of Gaussian distributions (DARG) \cite{wr8}.
\item \bfseries Riemannian manifold dimensionality reduction based methods: \mdseries Projection Metric Learning (PML) \cite{wr21}, SPD Manifold Learning (SPDML) based on affine-invariant metric (AIM) \cite{wr14} and stein divergence \cite{wr14} and Log-Euclidean Metric Learning (LEML) \cite{wr13}.
\item \bfseries Multiple statistical features based methods: \mdseries Localized Multi-Kernel Metric Learning (LMKML) \cite{wr37}, Hybrid Euclidean-and-Riemannian Metric Learning (HERML) \cite{wr31}, and Multiple Manifolds Metric Learning (MMML) \cite{mmml}.
\item \bfseries Deep learning based methods: \mdseries SPD Network (SPDNet) \cite{wr24} and Multi-Manifold Deep Metric Learning  (MMDML) \cite{wr36}.
\item \bfseries Other set model based method: \mdseries Discriminative Canonical Correlations (DCC) \cite{wr46} and Manifold-Manifold Distance (MMD) \cite{wr47}.
\end{itemize}

Here, we should point out the classification results of GDA, CDL, PML and LEML are carefully implemented by ourselves. As to other comparative methods, we adopt their source codes provided by the original authors to make experiments except for LMKML and MMDML. Since the source code of LMKML and MMDML have not been released, we use the classification rates that have already been achieved in \cite{wr36,wr32}. For fair comparison, the parameters of the comparative methods that we set in this paper are empirically tuned according to the original works. For CDL, KDA is used for discriminative subspace learning and the perturbation is set to $10^{-3}\times trace(C)$. In PML, the number of iterations and the value of the trade-off coefficient are set according to the original authors \cite{wr21}, and the target dimensionality of the generated new Grassmann manifold is determined by cross-validation. For GDA, we make use of the Projection Metric \cite{wr10} and its corresponding projection kernel. Moreover, the number of basis vectors for the subspace in GDA and GEDA are determined by cross-validation. In LEML, $\eta$ and $\zeta$ are the only two parameters need to be optimized, and we search their values in the range of $[0.1,1,10]$ and $[0.1:0.1:1]$. In SPDNet, the size of the transformation matrices are configured to $400\times200$, $200\times100$ and $100\times50$, respectively. Other parameters such as the number of epochs and the size of input data are set to $500$ and $400\times400$, while the learning rate and batch size are chosen by cross-validation. For RSR, we tune the value of $\lambda$ in the scope of $[0.0001,0.001,0.01,0.1]$. The two parameters $\upsilon_w$ and $\upsilon_b$ in SPDML-AIM and SPDML-Stein are searched by referring to \cite{wr14}, while the target dimensionality of the resulted new SPD manifold is set by cross-validation. In LMKML, the learning rate $\alpha$ is set as $10^{-6}$ and the bandwidth of Gaussian kernel is tuned by cross-validation. For HERML, we respectively tune the two parameters $\gamma$ and $\zeta$ in the scope of $[0.001,0.01,0.1,1,10,100,1000]$ and $[0.1:0.1:1]$. For DCC and MMD, we follow the default settings in \cite{wr46,wr47}.

\subsection{Video-based Face Recognition}
In this paper, the much challenging and widely used YouTube Celebrities (YTC) \cite{wr12,wr8,wr21,wr31} dataset is applied to the task of video-based face recognition. This dataset contains 1,910 video clips of 47 subjects that were collected from the website of YouTube. Each clip is comprised of hundreds of frames, most of which exhibit large variations in expression, illumination and pose. The number of image sets available for each subject is not fixed. Some sample face frames of this dataset are shown in Fig.2. Following the previous works \cite{wr12,wr8,wr21,wr31,wr36,wr13}, in our experiments we first reshape each face image into a $20\times20$ grayscale one and in order to eliminate the effects of lighting, histogram equalization is adopted for pre-processing. Then, we randomly select nine image sets in each subject with three for training and six for testing. Finally, we run ten times different combinations of gallery/probe and report the average recognition rates of different methods in Table 1.

 \begin{figure}[!t]
 \centering
 \includegraphics[scale=0.7]{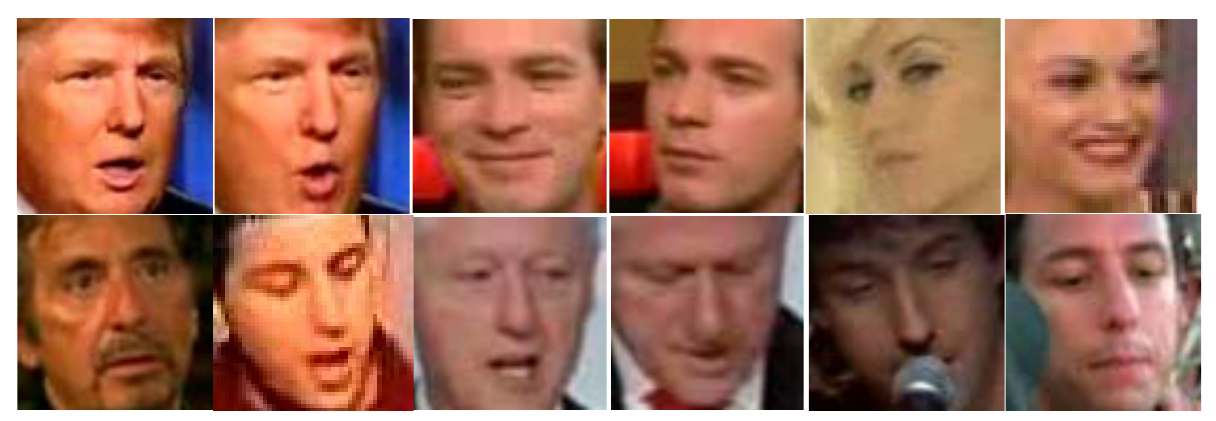}
 \caption{Face frames on YTC dataset}
 \label{fig:label}
\end{figure}

\begin{table}[!t]
 \newcommand{\tabincell}[2]{\begin{tabular}{@{}#1@{}}#2\end{tabular}}
 \centering
 \caption{\label{s_result} Average recognition rates (\%) of different methods on YTC dataset } 
 \rule[2.0cm]{0.0cm}{0.0pt}
 \begin{tabular}{c|c|c}
  \toprule 
   Method & Years & YTC \\ 
  \midrule 
   DCC \cite{wr46} & 2007 & 66.81\\
   MMD \cite{wr47} & 2008 & 65.30\\
  \hline
   GDA \cite{wr10} & 2008 & 65.78\\
   GEDA \cite{wr45} & 2011 & 66.37\\
   CDL \cite{wr12} & 2012 & 68.76\\
   RSR \cite{wr22} & 2012 & 72.77\\
  \hline
   LMKML \cite{wr37} & 2013 & 70.31\\
   HERML \cite{wr31} & 2015 & 71.28\\
   MMML \cite{mmml} & 2018 & 76.70\\
   \hline
   PML \cite{wr21} & 2015 & 67.62\\
   LEML \cite{wr13} & 2015 & 69.04\\
   SPDML-AIM \cite{wr14} & 2018 & 64.66\\
   SPDML-Stein \cite{wr14} & 2018 & 61.57\\
  \hline
   MMDML \cite{wr36} & 2015 & 69.81\\
   SPDNet \cite{wr24} & 2017 & 67.38\\
  \hline
  \bfseries Proposed \mdseries & & \bfseries 74.82 \mdseries \\
  \bottomrule 
 \end{tabular} 
\end{table}

According to the results listed in Table 1 we have some interesting observations. Firstly, the recognition rate of GDA is inferior to that of GEDA, which demonstrates the consideration of local structure of the data is helpful for us to extract more discriminative information when performing discriminant analysis on Grassmann manifold. Furthermore, it is intuitive to see the classification performance of both GDA and GEDA is lower than PML. The most fundamental reason is the process of directly performing dimensionality reduction on the Grassmann manifold can more faithfully characterize the geometry of the original set data than the Euclidean treatment. This reason can also be used to explain the difference in recognition rates between CDL and LEML. 

Secondly, compared with the state-of-the-art methods, the proposed algorithm shows better classification performance than them on this data. Among these results, we first want to make a comparison between SPMDL-AIM/SPDML-Stein and LEML. It can be found that LEML outperforms the former, which justifies the potential superiority of LEM based Riemannian manifold dimensionality reduction framework over AIM and Stein divergence based ones. An essential reason can be adduced: the inherent matrix-form of LEML can perserve more structural information of the space spanned by SPD matrices than the vector-form of SPDML-AIM/SPDML-Stein. Then, we want to discuss the performance of SPDNet. Obviously, it shows a relatively poor result than other SPD matrix learning methods, which may due to the limited number of training samples. 

Lastly, the comparison of classification performance between the proposed algorithm and LMKML and HERML is what we especially care about on this dataset. It is easy to observe that LMKML and HERML outperform most of the comparative methods in terms of recognition rate, which proves combining multiple statistical features of the original set data can yield more discriminative information than single model based methods. However, LMKML is absolutely surpassed by the proposed algorithm. As discussed before, the reason is the proposed algorithm attempts to learn data-specific kernel features instead of the unified one learned in LMKML, which can better preserve the original set data structure. For HERML, no distinction is made between different local regions in the produced kernel spaces in terms of discriminability, and therefore leads to weaker performance when compared with ours. 

\subsection{Set-based Object Categorization}
For set-based object categorization task, we conduct experiments on the ETH-80 dataset \cite{wr12,wr13}. This dataset consists of 8 categories: cows, cups, horses, dogs, tomatoes, cars, pears, and apples with 10 image sets per class. There are 41 images that collected from different perspectives in each image set and the size of each image is $256\times 256$. As shown in Fig.3, there are some sample images on ETH-80 dataset. In order to keep consistent with the original works \cite{wr12,wr13,wr37,wr36,wr32}, we first extract the gray-scale features for each original instance and adjust its size to $20\times 20$. Then, we randomly choose five objects in each category for training sets and the remaining five for query sets. Moreover, we randomly split this dataset into ten different pairs of training set and testing set, and the following table shows the average classification accuracies of different methods.

 \begin{figure}[!b]
 \centering
 \includegraphics[scale=0.7]{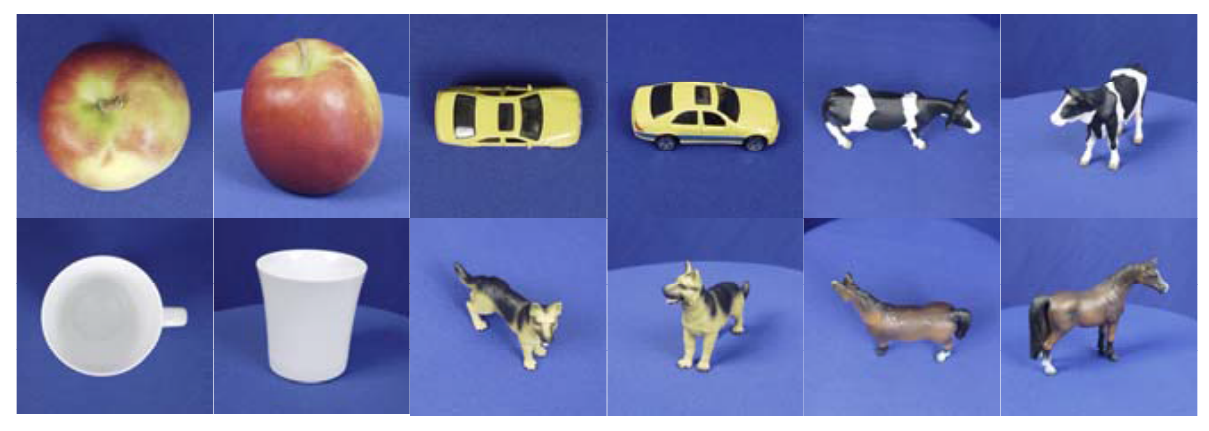}
 \caption{Examples on ETH-80 dataset}
 \label{fig:label}
\end{figure}

\begin{table}[!t]
 \newcommand{\tabincell}[2]{\begin{tabular}{@{}#1@{}}#2\end{tabular}}
 \centering
 \caption{\label{s_result} Average classification accuracies (\%) of different methods on ETH-80 dataset} 
 \rule[2.0cm]{0.0cm}{0.0pt}
 \begin{tabular}{c|c|c}
  \toprule 
   Method & Years & ETH-80 \\ 
  \midrule 
   DCC \cite{wr46} & 2007 & 90.75\\
   MMD \cite{wr47} & 2008 & 85.72\\
  \hline
   GDA \cite{wr10} & 2008 & 93.25\\
   GEDA \cite{wr45} & 2011 & 94.32\\
   CDL \cite{wr12} & 2012 & 93.75\\
   RSR \cite{wr22} & 2012 & 93.25\\
  \hline
   LMKML \cite{wr37} & 2013 & 92.50\\
   HERML \cite{wr31} & 2015 & 95.00\\
   MMML \cite{mmml} & 2018 & 95.00\\
   \hline
   PML \cite{wr21} & 2015 & 92.00\\
   LEML \cite{wr13} & 2015 & 92.25\\
   SPDML-AIM \cite{wr14} & 2018 & 90.75\\
   SPDML-Stein \cite{wr14} & 2018 & 90.50\\
  \hline
   MMDML \cite{wr36} & 2015 & 94.50\\
   SPDNet \cite{wr24} & 2017 & 86.25\\
  \hline
  \bfseries Proposed \mdseries & & \bfseries 96.25 \mdseries \\
  \bottomrule 
 \end{tabular} 
\end{table}

Among these classification results reported in Table 2, we summarize our observations in four aspects. The first is the classification performance of GEDA surpasses GDA, which further demonstrates the importance of exploiting the local structure of the set data. Then, it is interesting to observe the classification results generated by SPDML-AIM/SPDML-Stein are both inferior to that of LEML, which further justifies the matrix-form based SPD matrix feature learning is more effective than the vector-form based. Afterwards, we can intuitively find MMDML achieves better classification accuracy than SPDNet and other set-based methods. This mainly owes to the designed class-specific deep network in MMDML can extract more discriminative feature information for classification. However, SPDNet produces a relatively mediocre classification rate on this dataset, which further indicates the number of training sets plays a vital role in SPD manifold deep learning. Lastly, we also care about the classification performance of HERML and LMKML on this dataset. As can be found in Table 2, HERML achieves an impressive classification result and the classification performance of LMKML is also comparable. This again proves the complementary feature information provided by multiple statistics is useful to boost the image set classification performance. For the proposed algorithm, it yields a state-of-the-art classification result, which again demonstrates its effectiveness.

\subsection{Video-based Emotion Recognition}
For further evaluation, we apply the proposed algorithm to a much more difficult facial expression dataset for the emotion recognition task. This dataset is called Acted Facial Expression in Wild (AFEW) \cite{afew}, which depicts natural facial expressions in unconstrained environments and contains 1,345 video sequences of facial expressions collected from the movies with close to real world scenarios. Some examples on AFEW dataset are presented in Fig.4. To comply with the standard protocols of the Emotion Recognition in the Wild Challenge (EmotiW2014) \cite{afew}, we divide this dataset into three parts: training set, validation set and test set. Then, we follow \cite{afew,wr24} to split these training video sequences into 1,746 small clips for data augmentation. For the task of classifying each video sequence into one of the seven expression classes, we first resize each facial frame to an $20\times 20$ gray-scale image, then follow \cite{afew,wr24} to report the recognition results of different competitors on the validation set, dues to the ground truth of test set is not publicly available.

 \begin{figure}[!t]
 \centering
 \includegraphics[scale=0.52]{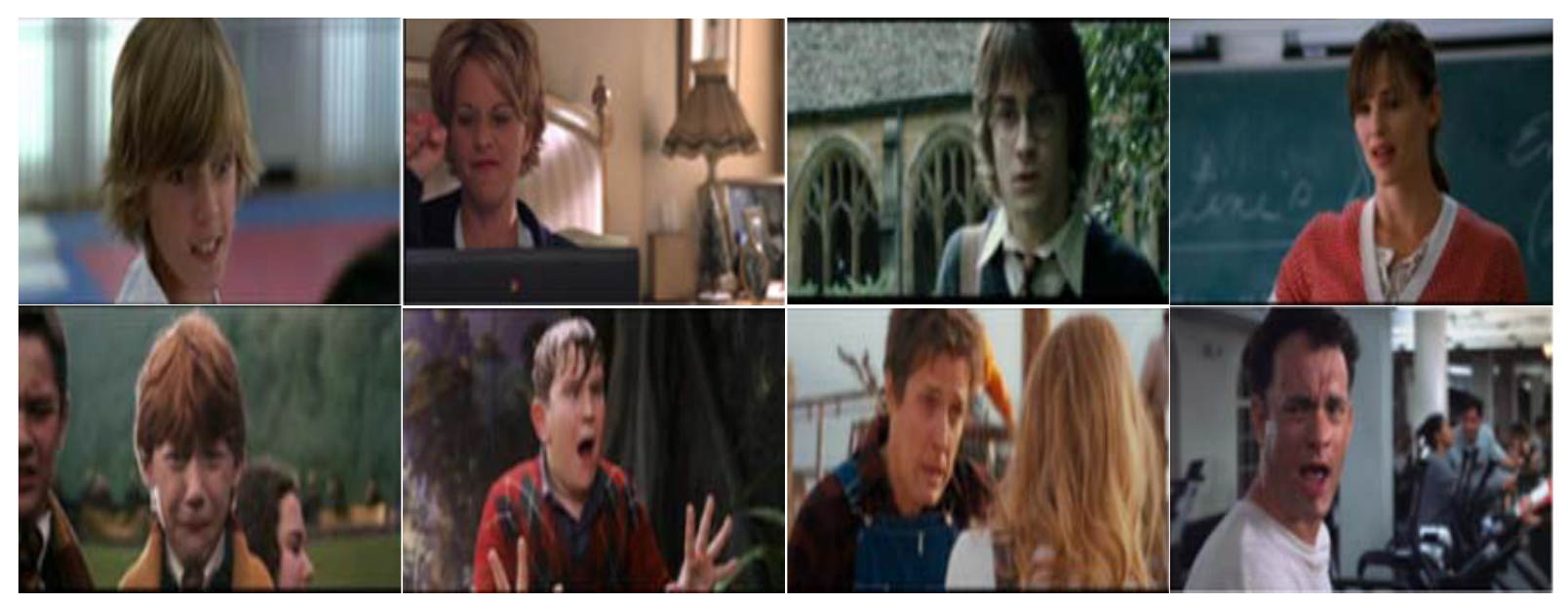}
 \caption{Emotion images on AFEW dataset}
 \label{fig:label}
\end{figure}

\begin{table}[!t]
 \newcommand{\tabincell}[2]{\begin{tabular}{@{}#1@{}}#2\end{tabular}}
 \centering
 \caption{\label{s_result} Emotion recognition results (\%) of different methods on AFEW dataset} 
 \rule[2.0cm]{0.0cm}{0.0pt}
 \begin{tabular}{c|c|c}
  \toprule 
   Method & Years & AFEW \\ 
  \midrule 
   DCC \cite{wr46} & 2007 & 25.78\\
  \hline
   GDA \cite{wr10} & 2008 & 29.11\\
   GEDA \cite{wr45} & 2011 & 29.45\\
   CDL \cite{wr12} & 2012 & 31.81\\
   RSR \cite{wr22} & 2012 & 27.49\\
  \hline
   HERML \cite{wr31} & 2015 & 32.14\\
   MMML \cite{mmml} & 2018 & 31.27\\
   \hline
   PML \cite{wr21} & 2015 & 28.98\\
   LEML \cite{wr13} & 2015 & 25.13\\
   SPDML-AIM \cite{wr14} & 2018 & 26.72\\
   SPDML-Stein \cite{wr14} & 2018 & 24.55\\
  \hline
   SPDNet \cite{wr24} & 2017 & 34.23\\
  \hline
  \bfseries Proposed \mdseries & & \bfseries 35.71 \mdseries \\
  \bottomrule 
 \end{tabular} 
\end{table}

According to the recognition results tabulated in Table 3, we can clearly find the performance of HERML exceeds most of the comparative methods on the task of emotion recognition. The reasons are arise from two aspects: 1) as discussed before, the multiple statistics can provide complementary feature information; 2) by jointly learning Euclidean-and-Riemannian metrics, more useful geometry information can be explored on this complicated video based dataset. Another interesting observation is the performance of LEML and SPDML-AIM/SPDML-Stein is more mediocre than CDL. This may be possible for that the linear transformation functions of LEML and SPDML-AIM/SPDML-Stein are learned on the non-linear manifold, which lose some ability to parse the structural information of complicated scenarios. Apparently, on this large-scale facial expression dataset, SPDNet shows its superiority on emotion recognition over other representative methods. Meanwhile, our method outperforms all the competitors, which demonstrates the integration of multiple Riemannian manifold-valued descriptors is qualified to improve the final classification accuracy.

\subsection{Dynamic Scene Classification}
Dynamic scene classification in an unconstrained setting is a fundamental and challenging task in computer vision. Recently, image set classification has provided a new direction to address this task. In this paper, we report the classification performance of our method on the MDSD \cite{wr35,wr49} dataset. This dataset is comprised of 13 different categories of dynamic scenes with each has 10 video sequences. As presented in Fig.5, there are some sample images on MDSD dataset. Due to the large intra-class variation in illumination, resolution, physical morphology and background, this classification seems very arduous. In our experiments, we follow the same protocols as introduced in the above experimental settings to preprocess each video frame, and use the seventy-thirty-ratio (STR) protocol, which typically builds gallery and probes by randomly selecting 7 videos for training set and the rest for query set in each category to test our method. Besides, we also conduct ten times different combinations of gallery/probe. The final average classification results are given in Table 4.

 \begin{figure}[!t]
 \centering
 \includegraphics[scale=0.52]{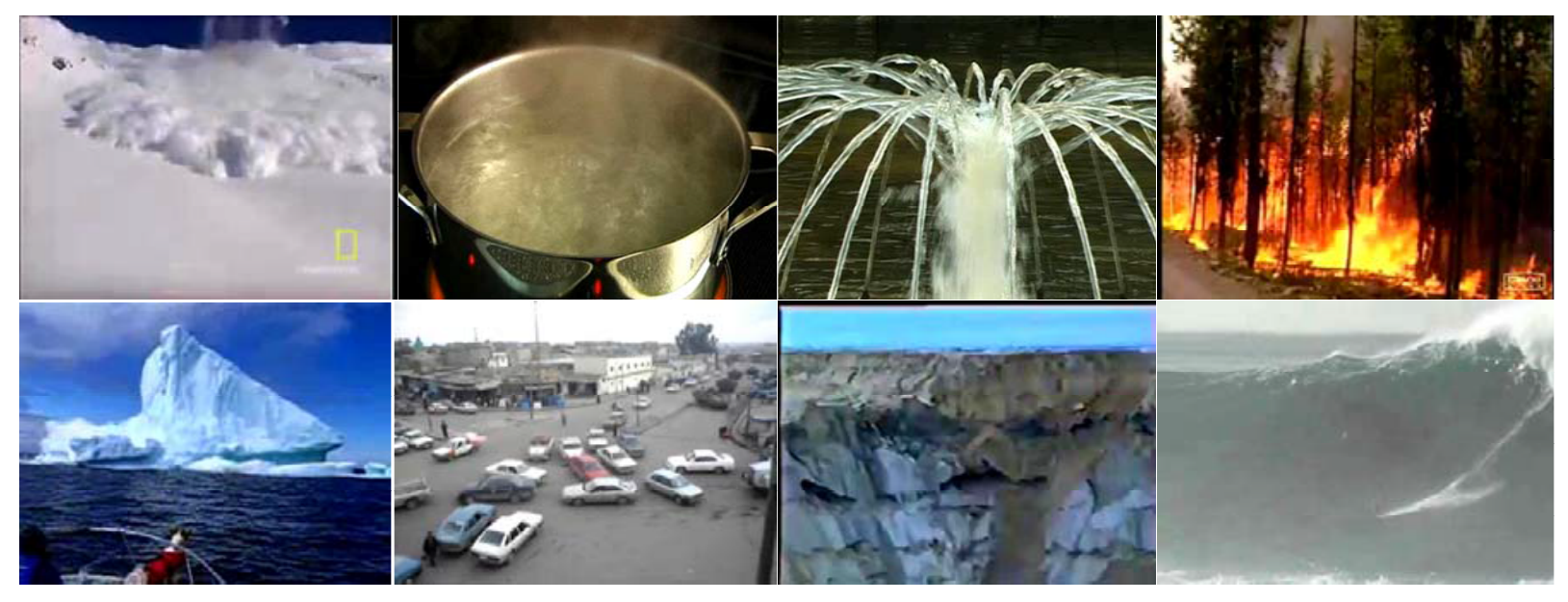}
 \caption{Dynamic scene images on MDSD dataset}
 \label{fig:label}
\end{figure}

\begin{table}[!t]
 \newcommand{\tabincell}[2]{\begin{tabular}{@{}#1@{}}#2\end{tabular}}
 \centering
 \caption{\label{s_result} Dynamic scene classification results (\%) of different methods on MDSD dataset} 
 \rule[2.0cm]{0.0cm}{0.0pt}
 \begin{tabular}{c|c|c}
  \toprule 
   Method & Years & MDSD \\ 
  \midrule 
   DCC \cite{wr46} & 2007 & 25.92\\
  \hline
   GDA \cite{wr10} & 2008 & 30.51\\
   GEDA \cite{wr45} & 2011 & 30.37\\
   CDL \cite{wr12} & 2012 & 30.51\\
   DARG \cite{wr8} & 2015 & 31.62\\
  \hline
   HERML \cite{wr31} & 2015 & 32.37\\
   MMML \cite{mmml} & 2018 & 31.95\\
   \hline
   PML \cite{wr21} & 2015 & 29.32\\
   LEML \cite{wr13} & 2015 & 29.74\\
   SPDML-AIM \cite{wr14} & 2018 & 31.10\\
   SPDML-Stein \cite{wr14} & 2018 & 29.81\\
  \hline
   SPDNet \cite{wr24} & 2017 & 32.05\\
  \hline
  \bfseries Proposed \mdseries & & \bfseries 36.67 \mdseries \\
  \bottomrule 
 \end{tabular} 
\end{table}

For evaluation, we compare the proposed algorithm with eleven state-of-the-art image set classification methods, as listed in Table 4. As can be seen in this table, our method produces a relatively good classification performance compared to others. However, the lower classification accuracies obtained by these methods intuitively illustrate this dynamic scene classification task is challenging. Then, we are interesting to see the performance of DARG surpasses other kernel based methods. The main reason is that the dissimilarity measurement between Gaussians in DARG is replaced by respectively measuring the dissimilarity between means and covariance matrices with Mahalanobis distance and LEM, which can extract more structural information for classification. For HERML, its classification performance is good on this dataset, and the same observations also can be found on the other three used datasets. This further justifies the effectiveness of jointly learning multiple statistics of image sets. Lastly, the achieved state-of-the-art classification results of the proposed method on all the used datasets verify its feasibility and utility.

\subsection{Ablation Study for Different Riemannian Manifold-valued Descriptors}

In previous experiments, the proposed algorithm has shown its superiority in image set classification over some representative set-based methods. Here, we further conduct experiments to observe the classification performance of each Riemannian manifold-valued descriptor incorporating with the proposed metric learning framework. Table 5 lists the classification results of them on the four used datasets, and some interesting observations can be summarized into two aspects. Firstly, for each used dataset, the classification results they have obtained are different. To be specific, on YTC and AFEW datasets the Grassmann manifold-valued descriptor achieves better classification performance than the other two, which may indicate the linear subspace is more effective in characterizing the structural information of the face image. On the contrary, the Gaussian distribution yields the best recognition rates on ETH-80 and MDSD datasets. There are two reasons may explain this: 1) most of the image sets in the two datasets conform to Gaussian distribution; 2) Gaussian model contains the first-order statistics and the second order statistics of the set data. Secondly, it is clear to see the performance of the proposed algorithm which simultaneously couples these three Riemannian manifold-valued descriptors with our multi-kernel metric learning framework outperforms the way of separately, which further justifies the complementarity of these three descriptors in set data modeling.

\begin{table}[!t]
 \centering
 \caption{\label{s_result} Average classification results (\%) of different Riemannian manifold-valued descriptors on the four used dataset} 
 \rule[1.0cm]{0cm}{0pt}
 \begin{tabular}{c|c|c|c|c}
  \toprule 
   Method & YTC & ETH-80 & AFEW & MDSD \\ 
  \midrule 
   Covariance Matrix & 72.22 & 95.31 & 32.43 & 31.03\\
   Linear Subspace & 73.58 & 95.63 & 34.05 & 35.13\\
   Gaussian Distribution & 70.57 & 95.94 & 31.08 &  36.15\\
  \hline
  \bfseries Proposed \mdseries & \bfseries 74.82 \mdseries & \bfseries 96.25  \mdseries  & \bfseries 35.71 \mdseries & \bfseries 36.67 \mdseries \\
  \bottomrule 
 \end{tabular} 
\end{table}

\subsection{Ablation Study for Convergence Behavior} 
 As discussed in Section 4.2.2, we expect to study the transformation matrix $E$ but have to infer $\delta_q$ and $\rho_q$ simultaneously. Hence, we use an iterative manner to solve this problem. To optimize $E$, we follow \cite{wr44} to directly solve the trace ratio problem defined in Eq.27, and for $\delta_q$ and $\rho_q$ we utilize gradient ascent method. Although, it is hard for us to provide a systematic theoretical proof of convergence behavior of this optimization problem, we find after several iterations the objective function Eq.32 can reach to a stable value, which is confirmed experimentally. Fig.6 and Fig.7 were respectively obtained using the AFEW dataset and YTC dataset, and it is intuitively observe that with the increase of the number of iterations, the value of the objective function tends to steadily fluctuate within a very small range. Furthermore, we also increase the number of iterations to 40 to see the current values of the objective function on the two datasets, which are 0.8398 and 0.9207 respectively. This demonstrates our algorithm can achieve a stable classification performance with more iterations.
 
 \begin{figure}[!t]
 \centering
 \includegraphics[scale=0.52]{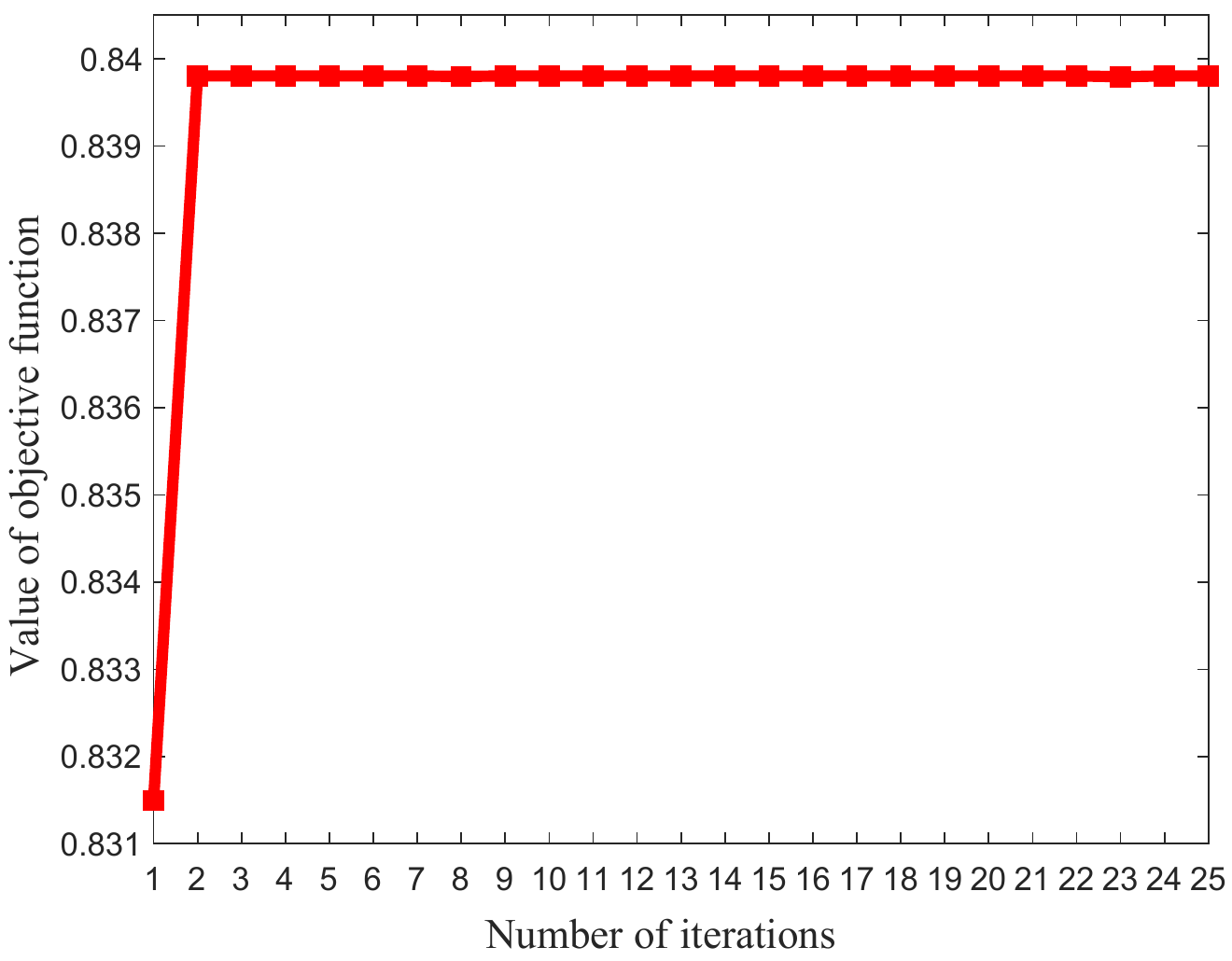}
 \caption{Convergence behavior of the proposed algorithm on AFEW dataset}
 \label{fig:label}
\end{figure}
 
 \begin{figure}[!t]
 \centering
 \includegraphics[scale=0.52]{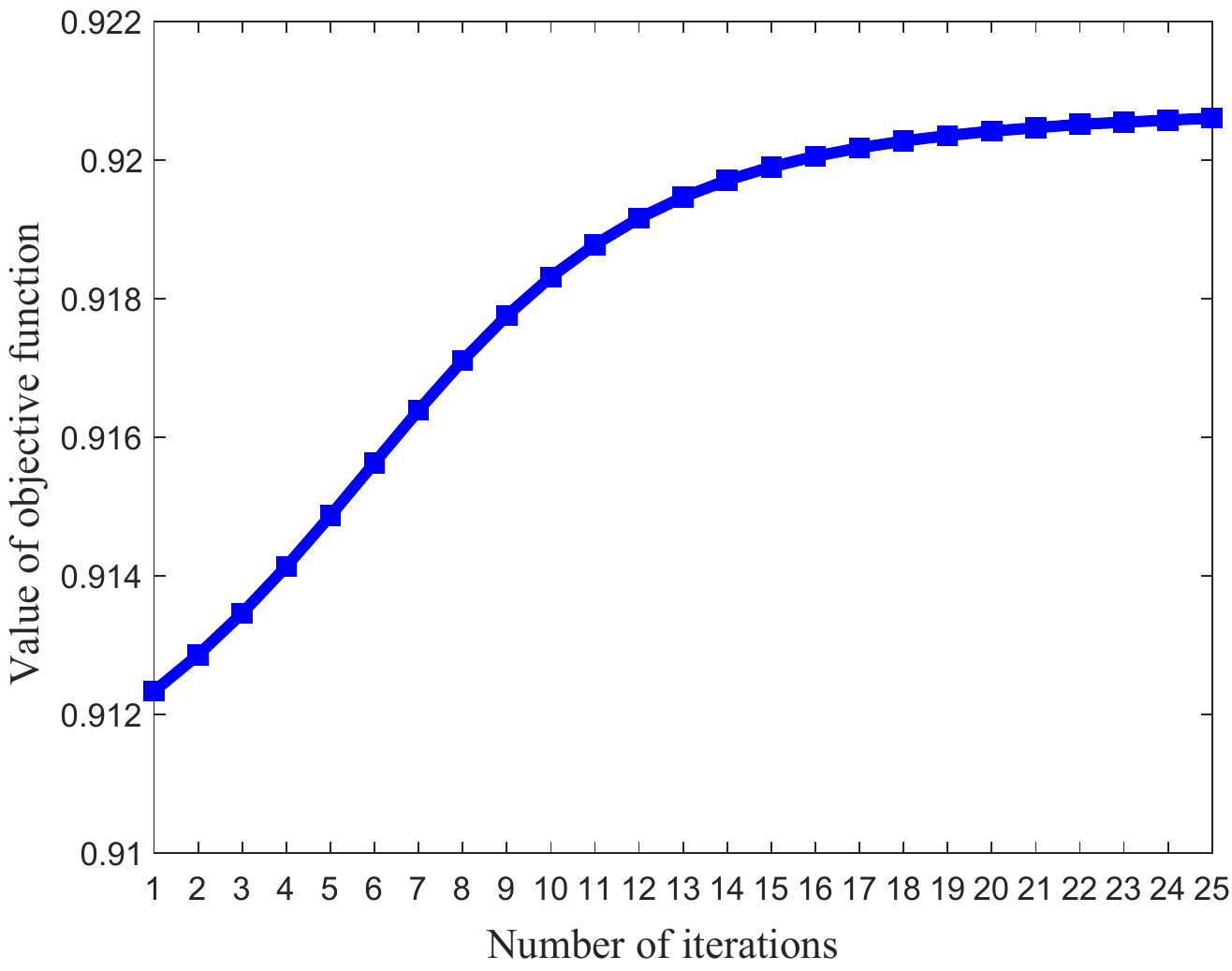}
 \caption{Convergence behavior of the proposed algorithm on YTC dataset}
 \label{fig:label}
\end{figure}

\subsection{Parameter Discussion}
 
 According to the description in Section 4, we can see the resulting high dimensional heterogeneous and complementary Riemannian manifold-valued features are fused into a $d_w$-dimensional Euclidean space under the proposed multi-kernel metric learning framework. Since more useful and more compact feature representations often reside in a lower dimensional feature space, it is indispensable to find a desirable value of $d_w$. Therefore, we make experiments on YTC and MDSD datasets to compare the impact of different $d_w$ on the final classification results of our method. The experimental results are presented in Fig.8 and Fig.9, respectively. From Fig.8, we can intuitively see it achieves a top classification result when the value of $d_w$ is 25. Moreover, we let $d_w$ reach to its maximum value on MDSD dataset, which is 91, but the produced 5.13\% classification accuracy is very lower than other cases. From Fig.9, we can easily find 70 is the best value of $d_w$ on YTC dataset. Furthermore, we also increase $d_w$ to 141, its maximum value on this dataset, and the generated 72.22\% recognition rate is somewhat lower. 
 
 With the above observations in mind, we can see there are two reasons that can explain the varying tendency of the curves depicted in Fig.8 and Fig.9. The first is when the values of $d_w$ are lower, the learned insufficient discriminative feature information is unable to make effective distinctions between some overlapping samples, which may bring about lower classification accuracies. The second is when the values of $d_w$ are higher, some redundant information cannot be effectively filted out from the extracted efficient features, which may also lead to undesirable recognition rates. Besides, on AFEW and ETH-80 datasets, the best values of $d_w$ are configured as 70 and 8, respectively.
 
 \begin{figure}[!t]
 \centering
 \includegraphics[scale=0.523]{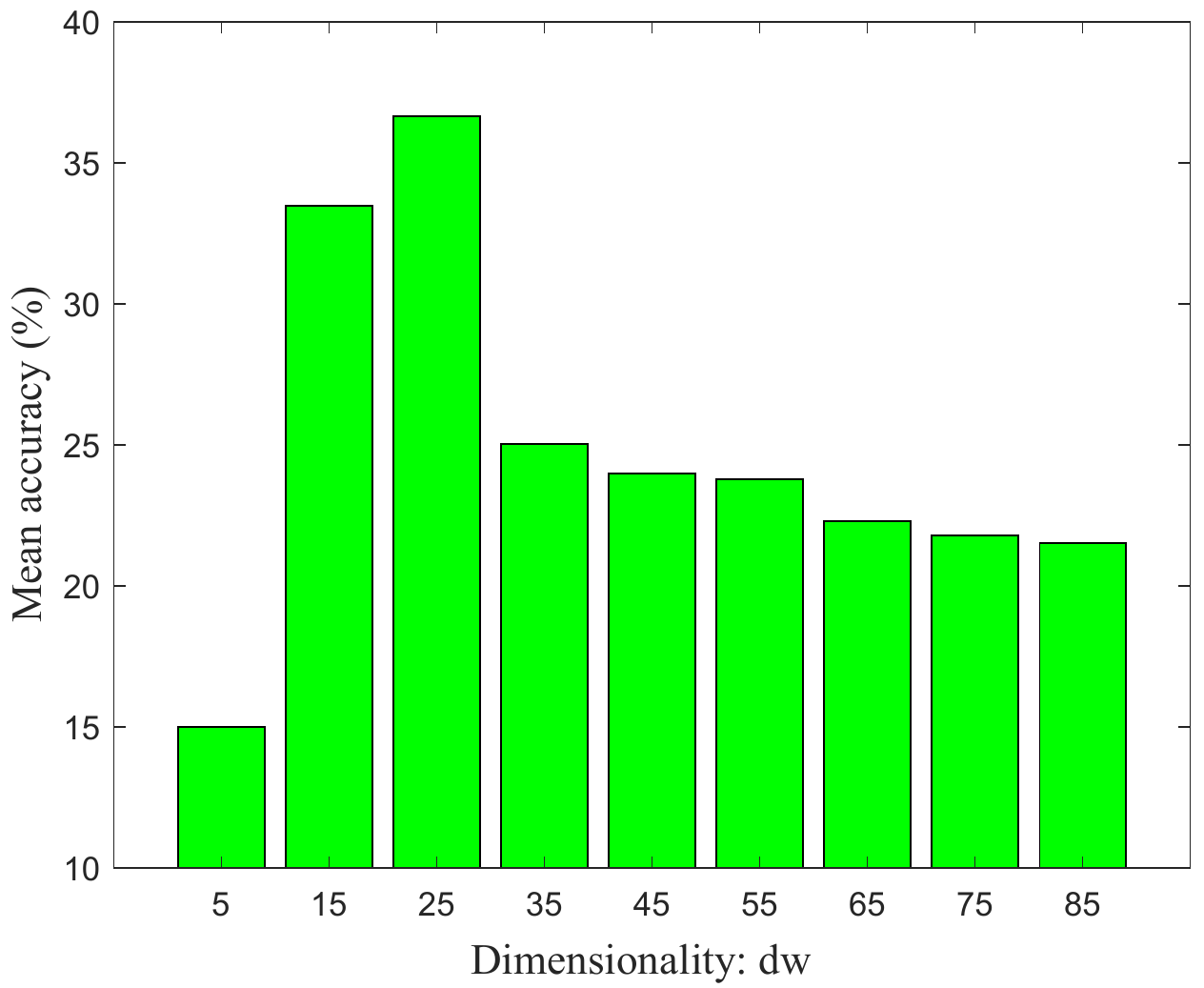}
 \caption{Average recognition rates (\%) produced by our method versus different $d_w$ on MDSD dataset}
 \label{fig:label}
\end{figure}
 
 \begin{figure}[!t]
 \centering
 \includegraphics[scale=0.523]{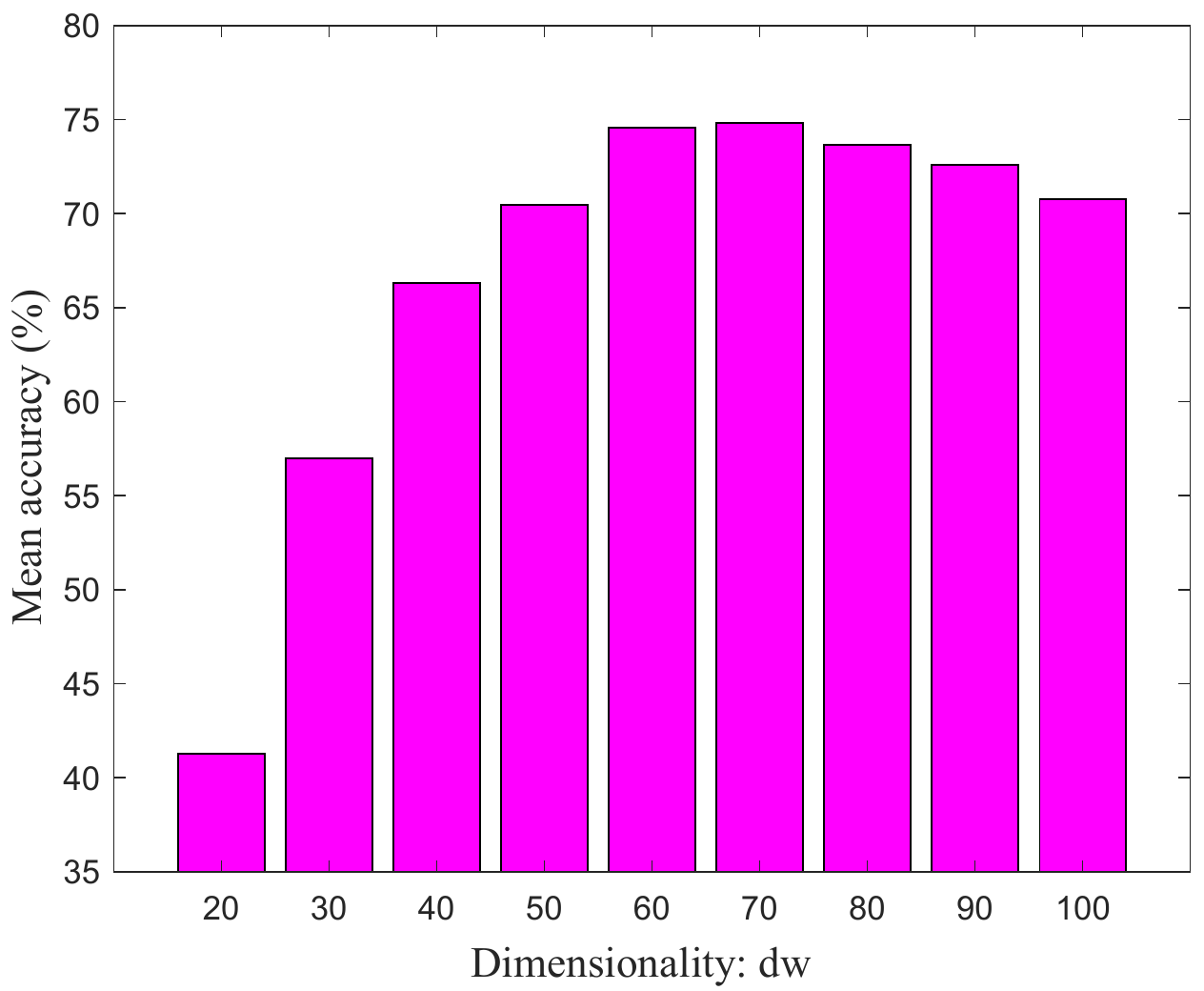}
 \caption{Average recognition rates (\%) produced by our method versus different $d_w$ on YTC dataset}
 \label{fig:label}
\end{figure}

\section{Conclusions}

In this paper, we propose a novel image set classification algorithm which fuses multiple Riemannian manifold-valued features of image sets with a designed multi-kernel metric learning framework. This proposed algorithm has been assessed on four image set classification tasks: video-based face recognition, set-based object categorization, video-based emotion recognition and dynamic scene classification respectively. Extensive experimental results computed on four video-based datasets demonstrate its superiority over some representative image set classification methods. Besides, the comparison between each single Riemannian manifold-valued descriptor and their combination justifies their complementarity in encoding the set data, and their fusion is beneficial to improve the classification performance on video-based set data.

Since the temporal order is an important factor in describing frames in video, we plan to integrate it into our proposed framework and hope this can help to improve its discriminatory ability on some complicated classification tasks. For future work, another possible direction is to investigate other metric learning methods to fuse the heterogeneous and complementary features. Finally, we would like to transfer some popular Euclidean deep learning architectures into Riemannian manifold for better recognition on large-scale video-based datasets.

\ifCLASSOPTIONcaptionsoff
  \newpage
\fi

\end{document}